%% file: main.tex
\documentclass[lettersize,journal]{IEEEtran}
\usepackage{amsmath,amsfonts}
\usepackage{algorithmic}
\usepackage{algorithm}
\usepackage{array}
\usepackage[caption=false,font=normalsize,labelfont=sf,textfont=sf]{subfig}
\usepackage{textcomp}
\usepackage{stfloats}
\usepackage{url}
\usepackage{verbatim}
\usepackage{graphicx}
\usepackage{cite}
\usepackage{hyperref}
\newtheorem{definition}{Definition}
\usepackage{bbding}
\usepackage{enumitem}
\usepackage{bbm}
\usepackage{booktabs}
\usepackage[export]{adjustbox}
\usepackage{xspace}
\usepackage{multirow}

\usepackage{color}
\usepackage{colortbl}
\definecolor{Gray}{gray}{0.9}
\newcommand{\hide}[1]{} 

\newcommand*{\vpara}[1]{\vspace{0.1in}\noindent\textbf{#1}}

\newcommand{\model}[0]{PGT\xspace}

\begin{document}

\title{Generalizing Graph Transformers Across Diverse Graphs and Tasks via Pre-training}


\author{Yufei He, Zhenyu Hou, Yukuo Cen, Jun Hu, Feng He, Xu Cheng, Jie Tang,~\IEEEmembership{Fellow,~IEEE,} Bryan Hooi* 

\IEEEcompsocitemizethanks{
\IEEEcompsocthanksitem Yufei He, Jun Hu, and Bryan Hooi are with the National University of Singapore.
\IEEEcompsocthanksitem Zhenyu Hou, Xu Cheng, and Jie Tang are with the Department of Computer Science, Tsinghua University.
\IEEEcompsocthanksitem Feng He is with Tencent.
\IEEEcompsocthanksitem Yukuo Cen is with Zhipu AI.
\IEEEcompsocthanksitem *Corresponding author: Bryan Hooi (\href{mailto:dcsbhk@nus.edu.sg}{dcsbhk@nus.edu.sg}). 
}
}

\markboth{IEEE Transactions on Knowledge and Data Engineering
}%
{Shell \MakeLowercase{\textit{et al.}}: A Sample Article Using IEEEtran.cls for IEEE Journals}


\maketitle

\input{0.1.abstract}


\input{1.introduction}

\input{2.related}
\input{3.method}
\input{4.experiments}

\input{5.conclusion}

\section*{Acknowledgments}
This research is supported by the Ministry of Education, Singapore, under the Academic Research Fund Tier 2 (FY2025) (Grant MOE-T2EP20124-0009)



 

\bibliographystyle{IEEEtran}
\bibliography{reference}



 






\end{document}

%% file: 0.1.abstract.tex
\begin{abstract}
Graph pre-training has been concentrated on graph-level tasks involving small graphs (e.g., molecular graphs) or learning node representations on a fixed graph. Extending graph pre-trained models to web-scale graphs with billions of nodes in industrial scenarios, while avoiding negative transfer across graphs or tasks, remains a challenge. We aim to develop a general graph pre-trained model with inductive ability that can make predictions for unseen new nodes and even new graphs. In this work, we introduce a scalable transformer-based graph pre-training framework called \model (Pre-trained Graph Transformer)\footnote[1]{The code is available at \url{https://github.com/yf-he/PGT}.}. Based on the masked autoencoder architecture, we design two pre-training tasks: one for reconstructing node features and the other for reconstructing local structures. Unlike the original autoencoder architecture where the pre-trained decoder is discarded, we propose a novel strategy that utilizes the decoder for feature augmentation. 
Our framework, tested on the publicly available ogbn-papers100M dataset with 111 million nodes and 1.6 billion edges, achieves state-of-the-art performance, showcasing scalability and efficiency.
We have deployed our framework on Tencent’s online game data, confirming its capability to pre-train on real-world graphs with over 540 million nodes and 12 billion edges and to generalize effectively across diverse static and dynamic downstream tasks.
\end{abstract}

%% file: 1.introduction.tex
\section{Introduction}

\IEEEPARstart{G}{raph} data, characterized by its intricate network of interconnected nodes and edges, pervades a multitude of domains in the real world. This pervasive presence arises from the inherent relational structures observed in various systems, ranging from social networks~\cite{liao2018attributed,sui2024fidelis} and transportation networks~\cite{guo2019attention} to biological networks~\cite{ying2021transformers} and financial systems~\cite{wang2021review}. 
Graph Neural Networks (GNNs) have been a dominant model architecture in modeling complex relationships within graph data~\cite{kipf2016semi,velivckovic2017graph,hamilton2017inductive}, making them versatile tools for addressing real-world challenges and fostering innovation~\cite{ying2018graph,masters2022gps++,allamanis2022graph,DBLP:journals/tkde/HuHQFX24,DBLP:conf/aaai/0016HHW25,sui2024can,liu2025efficient,liu2025guardreasoner,he2025enabling,he2025evotest,chen2025can,chen2025robustness,wang2025safety,he2025evaluating,zhang2021scr,he2022sgkd,sui2025meta}.

\begin{figure}[t]
    \centering
    \includegraphics[width=0.7\linewidth]{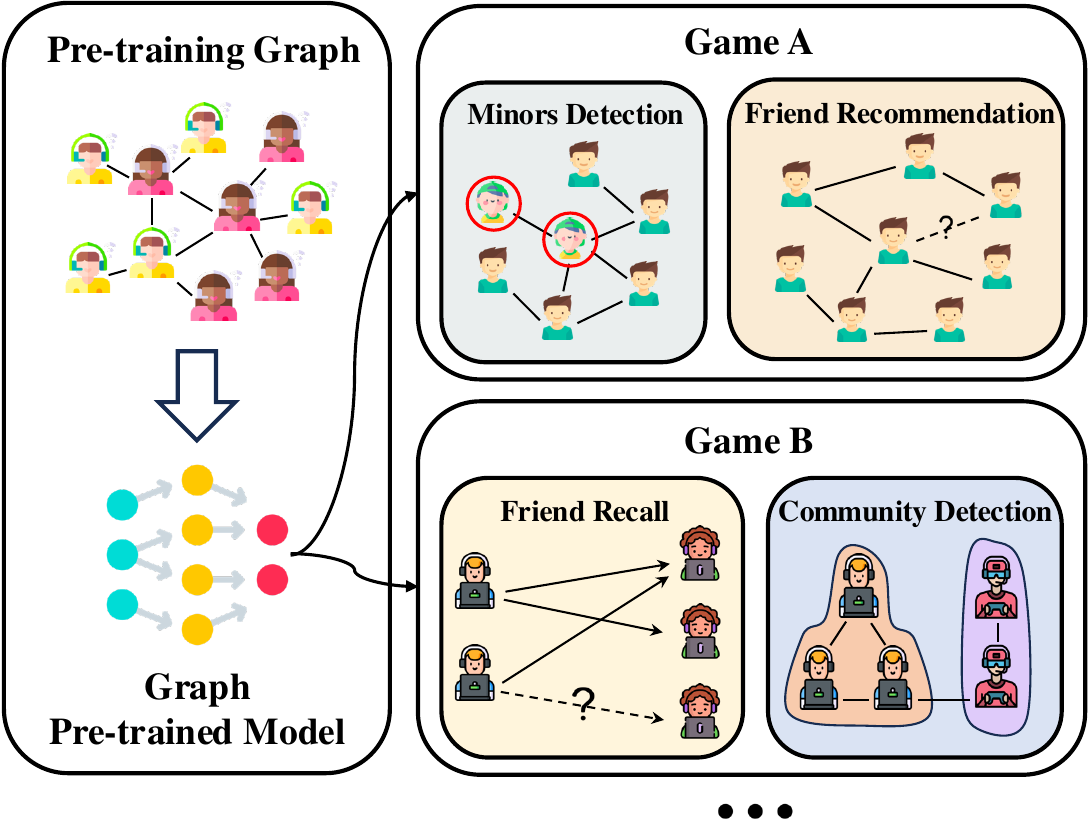}
    \vspace{-4mm}
    \caption{Illustrations of graph pre-training in the online gaming industry.}
    \label{fig:fig1}
    \vspace{-5mm}
\end{figure}

Online gaming, as a domain of interest in network science, presents a rich source of data due to its inherent graph-like structure. 
Advanced graph learning techniques are increasingly recognized as an important means for enhancing data analysis in the online gaming industry~\cite{yang2022large,cheng2025lps}.

\vpara{Challenges.} While there have been some initial attempts to apply graph neural networks in real-world industrial settings~\cite{wang2019deep,zhang2023inferturbo}, deploying GNNs in products and bridging the gap between academia and industry remains a long journey. 

Firstly, message-passing GNNs, constrained by strong inductive biases, do not consistently perform well on real-world industrial graph data. Graph data in the industry may unavoidably contain structural noise, causing GNNs, which aggregate information from pre-defined neighbors, to learn indistinct representations~\cite{fox2019robust}. 
An example is heterophilic graphs that do not adhere to the homophily assumption, where node features and labels significantly differ from those of their neighbors~\cite{zheng2022graph}. Designing an appropriate backbone network for online gaming networks is a significant challenge.

Secondly, existing graph learning methods often struggle to meet the requirements of scalability and efficiency in inference posed by online gaming businesses. 
The size of real-world online gaming networks in industry is typically 5-6 times larger than the largest publicly available datasets in academia. 
In addition, existing scalable GNNs tend to prioritize training over inference~\cite{chiang2019cluster,zeng2019graphsaint}. 
This leads to a need to design novel sampling methods to meet the business demands for high-throughput and low-latency model inference.

Thirdly, as illustrated in Fig.~\ref{fig:fig1}, differing from most existing works that focus on optimizing models for a specific task on a fixed graph, the online gaming industry often involves highly diverse graphs and downstream tasks. 
For example, friend relationships in various games can be modeled as different graphs, and within a single game, there exist various prediction tasks like user classification and friend recommendations. 
Training models from scratch for each task on each graph can result in a significant waste of time and computational resources. 
Given the great success of pre-trained models in computer vision~\cite{radford2021learning} and natural language processing~\cite{brown2020language}, intuitively, we aim for a graph pre-trained model that serves as a general model that can effectively generalize across different graphs and tasks.
However, in practice, due to the high diversity of graph data, graph pre-trained models often exhibit \textit{negative transfer}, which hinders their performance~\cite{hu2019strategies,qiu2020gcc,cao2023pre,he2025unigraph,he2025unigraph2}.

\vpara{Contributions.}
In this work, we present \model, a new graph pre-training framework with transformers. 
Instead of message-passing GNNs, we adopt the Transformer model~\cite{vaswani2017attention} as the backbone, which has demonstrated its superiority in various domains~\cite{devlin2018bert,dosovitskiy2020image}. 
Graph transformers disregard pre-defined graph structures and learn a soft, fully connected graph structure bias across all nodes. 
This approach can mitigate the graph structural noise and heterophily present in industrial data. 

Applying the Transformer architecture to web-scale graphs is not a trivial endeavor, primarily due to the computational and memory costs induced by the quadratic complexity of the self-attention mechanism.
We employ the classic Personalized PageRank (PPR) to sample contextual node sequences for each seed node. 
This also decouples the sampling of labeled and auxiliary nodes, aligning well with industrial scenarios where labeled nodes constitute an extremely small fraction of the entire graph.
In contrast to expensive random data accesses on the entire graph~\cite{chiang2019cluster,zeng2019graphsaint}, our method merely accesses the corresponding node sequences. 
Experiments show our method accelerates inference by up to 12.9x while achieving better performance compared to previous methods.

In the online gaming industry, due to the growing complexity and scale of games themselves and the interactions between players, entirely new graphs will continue to emerge as the business grows.
The objective of graph pre-trained models is to generalize to unseen new nodes, edges, and even entirely novel graphs. 
The scalable graph transformer model we propose effectively mitigates the issue of structural distribution discrepancies between graphs (such as degree distribution) by implicitly learning the local structure of the graph from contextual sequences. 
Based on the Masked Graph Modeling (MGM) paradigm, we design two distinct pre-training tasks: one at the node-level for node feature reconstruction, and the other at the edge-level for local structural reconstruction. 
In the inference phase, we introduce a novel feature augmentation technique. 
In contrast to the conventional Masked Autoencoder (MAE) architecture, which discards the decoder after pre-training~\cite{he2022masked}, we utilize the pre-trained decoder to generate reconstructed node features. 
These reconstructed features are then averaged with the original features and used as input for the encoder during inference. 

We conduct extensive experiments on Tencent's online gaming data to demonstrate the performance, efficiency, and transferability of \model. We perform pre-training on a Tencent gaming user friendship graph comprising 540 million nodes and 12 billion edges. Subsequently, the pre-trained model is applied to two downstream tasks across four different games. 
Additionally, we experiment with publicly available benchmarks. The results on Tencent's online gaming dataset and public datasets suggest that, when compared to existing graph self-supervised learning algorithms, the \model framework is capable of achieving state-of-the-art performance.

Also, graph pre-training extends beyond online gaming applications and addresses fundamental challenges in graph learning across various domains. 
In finance, graph pre-trained models can enhance fraud detection and credit risk assessment by capturing complex relationships between transactions and entities despite heterogeneous graph structures. 
In healthcare, these models can improve patient outcome prediction and drug discovery by transferring knowledge across different medical data graphs with varied structures. 
The scalability and cross-graph generalization capabilities of our approach are particularly valuable in these domains, where graph data often exhibits significant structural noise and heterophily. 
Through our comprehensive experiments on both gaming data and public academic datasets spanning different domains, we demonstrate the broad applicability of our framework across diverse real-world settings.

Overall, our work makes the following four contributions:
\begin{itemize}[leftmargin=*,itemsep=0pt,parsep=0.2em,topsep=0.3em,partopsep=0.3em]
    \item We propose the Pre-trained Graph Transformer (\model) framework, which employs two pre-training tasks to enable the model to learn transferable node features and graph structural patterns. 
    \item We propose a novel strategy of using the pre-trained decoder for feature augmentation, which further elevates the performance of the pre-trained model on downstream tasks.
    \item We conduct extensive experiments on both industrial and public datasets, demonstrating that PGT can effectively generalize to diverse graphs and tasks. Crucially, we show that a simple adaptation of our statically pre-trained model can serve as a powerful backbone for dynamic graph tasks, outperforming specialized dynamic GNNs trained from scratch. Our empirical findings substantiate that PGT yields superior results compared to other graph self-supervised algorithms.
\end{itemize}

%% file: 2.related.tex
\section{Related work}
\subsection{Scalable Graph Neural Networks}
Standard Graph Neural Networks (GNNs)~\cite{hamilton2017inductive,kipf2016semi,velivckovic2017graph,xu2018powerful} struggle with large-scale graphs due to memory constraints. A primary solution is sampling, including node-level sampling (GraphSAGE~\cite{hamilton2017inductive}) and sub-graph sampling (Cluster-GCN~\cite{chiang2019cluster}, GraphSAINT~\cite{zeng2019graphsaint}). While most work focuses on training, methods like ShaDow-GNN~\cite{zeng2021decoupling} also address scalable inference. A complementary approach simplifies the GNN architecture itself. Linear GNNs like SGC~\cite{wu2019simplifying} remove non-linearities, collapsing propagation into a single matrix operation. Other graph reduction methods use techniques like coarsening or sparsification to shrink the graph~\cite{ma2025acceleration}. However, these simplification techniques can lose crucial structural information on large, heterogeneous industrial graphs.

\subsection{Graph Transformers}
Inspired by their success in NLP~\cite{vaswani2017attention} and computer vision~\cite{dosovitskiy2020image}, Transformers have been adapted for graph learning. Early Graph Transformers focused on small graphs, incorporating structural information via positional encodings, such as Laplacian eigenvectors (GT~\cite{dwivedi2020generalization}, LiteGT~\cite{chen2021litegt}, SAN~\cite{chen2022structure}) or centrality measures (Graphormer~\cite{ying2021transformers}). The quadratic complexity of self-attention hinders their application to large graphs. To address this, scalable Graph Transformers employ techniques like neighbor sampling (Gophormer~\cite{zhao2021gophormer}, ANS-GT~\cite{zhang2022hierarchical}) or efficient attention approximations like kernelized operators (NodeFormer~\cite{wu2022nodeformer}).

\subsection{Pre-training on Graphs.}
Pre-training has become a key paradigm for GNNs. Initial work used supervised pre-training on domain-specific data~\cite{hu2019strategies}. More recent efforts focus on self-supervised learning, which broadly falls into two categories. Contrastive methods learn by maximizing agreement between positive views, using cross-scale contrast (DGI~\cite{velivckovic2018deep}, MVGRL~\cite{hassani2020contrastive}) or, inspired by BYOL~\cite{grill2020bootstrap}, avoiding negative samples via self-distillation (BGRL~\cite{thakoor2021bootstrapped}). Generative methods use an encoder-decoder framework to reconstruct masked node features (MGAE~\cite{wang2017mgae}, GraphMAE~\cite{hou2022graphmae}, GraphMAE2~\cite{hou2023graphmae2}, GMAE~\cite{zhang2022graph}) or graph structure (GAE~\cite{kipf2016variational}, VGAE~\cite{kipf2016variational}, MaskGAE~\cite{li2022maskgae}). However, the scalability and cross-graph transferability of these methods for web-scale industrial graphs with varying structures remain open challenges~\cite{qiu2020gcc,cao2023pre}.

\subsection{Learning on Dynamic Graphs}
A related but distinct challenge is learning on dynamic graphs, where the graph structure and features evolve over time. While our work focuses on pre-training for generalization across diverse \textit{static} graphs, dynamic graph representation learning methods are specifically designed to capture temporal patterns. These models, such as EvolveGCN~\cite{pareja2020evolvegcn} and TGN~\cite{rossi2020temporal}, often use mechanisms like recurrent networks or memory modules to efficiently update representations as the graph changes. DGCN~\cite{gao2022novel} utilizes an LSTM to update the weight parameters of a GCN, aiming to capture global structural information across temporal snapshots. Other notable approaches include VGRNN~\cite{hajiramezanali2019variational}, a variational generative model. Although some of our industrial tasks involve temporal snapshots, our core contribution addresses the orthogonal problem of negative transfer between structurally different graphs, rather than modeling temporal evolution within a single graph. 

%% file: 3.method.tex
\section{Preliminaries}

We begin by introducing essential concepts and notations. 

\vpara{Graph.}
A graph is denoted as $\mathcal{G} = (\mathcal{V}, \mathcal{E})$, where $\mathcal{V}$ represents the node set, and $\mathcal{E} \subseteq \mathcal{V} \times \mathcal{V}$ is the edge set. These connections between nodes can also be expressed using an adjacency matrix $\mathbf{A} \in \mathbb{R}^{N \times N}$, where $N$ is the number of nodes, i.e., $|\mathcal{V}|=N$. The element at the $i$-th row and $j$-th column of $\mathbf{A}$ is denoted as $\mathbf{A}_{ij}$, equaling $1$ if nodes $v_i$ and $v_j$ are connected, otherwise $\mathbf{A}_{ij} = 0$. Each node $v_i \in \mathcal{V}$ is associated with a feature vector $\mathbf{x}_i \in \mathbb{R}^{d}$, where $d$ is the feature dimension. The feature information for all nodes is represented as a matrix $\mathbf{X} \in \mathbb{R}^{N \times d}$, with the $i$-th row of $\mathbf{X}$ corresponding to the features of node $v_i$.

\subsection{Graph Pre-training: Challenges}
The goal of graph pre-training is to learn a model on a graph domain that generalizes to unseen graphs and benefits various downstream tasks. Achieving this presents two primary challenges. First, the model's backbone must have sufficient capacity to learn transferable knowledge while remaining scalable for both training and inference. Many existing scalable GNNs are shallow and focus primarily on training efficiency, often overlooking inference~\cite{chiang2019cluster,zeng2019graphsaint}. Second, and more critically, pre-trained models often suffer from \textit{negative transfer} when applied across graphs with diverse structures~\cite{hu2019strategies,qiu2020gcc}. For instance, social networks can vary from dense, bidirectional graphs (like Facebook) to sparser, star-shaped structures (like Twitter). The strong inductive biases of GNNs make it difficult to generalize across such divergent local structures.

Existing graph neural networks exhibit specific limitations when confronted with diverse graph structures, primarily due to their rigid message-passing mechanisms. 
Message-passing GNNs inherently assume that connected nodes share similar features or labels (homophily), which proves problematic in heterophilic graphs where connected nodes tend to have dissimilar characteristics. 
For instance, in online gaming networks, users with opposing play styles or from different demographic groups often connect, creating heterophilic patterns that conventional GNNs struggle to model. 
Additionally, these models are highly sensitive to structural noise—common in industrial graphs due to frequent changes in user relationships, spam connections, or adversarial behaviors. For example, in online gaming networks, temporary team formations or automated friend requests can introduce misleading structural patterns. 
This issue is particularly severe in cross-graph transfer scenarios, where pre-trained GNNs fail to generalize because they overly adapt to the specific structural patterns of the training graph. Empirical studies show that when deployed on graphs with different structural characteristics (e.g., degree distributions, clustering coefficients, or assortativity), pre-trained GNNs often perform worse than randomly initialized models, demonstrating negative transfer effects.

\subsection{Negative Transfer in Graph Pre-training} 
\begin{definition}[Negative Transfer in Graph Pre-training]
    Let $M_{pre}$ be a graph model pre-trained on source graph $G_s$, and $M_{rand}$ be the same model architecture but with random initialization. Negative transfer occurs when, after applying both models to a downstream task on target graph $G_t$ (either through linear probing or fine-tuning), the performance of $M_{pre}$ is worse than $M_{rand}$, i.e., $\text{Perf}(M_{pre}, G_t) < \text{Perf}(M_{rand}, G_t)$, where $\text{Perf}(\cdot,\cdot)$ is a task-specific metric.
\end{definition}

\section{The \model Framework}

\begin{figure*}[!t]
\vspace{-7mm}
    \centering
    \includegraphics[width=0.8\linewidth]{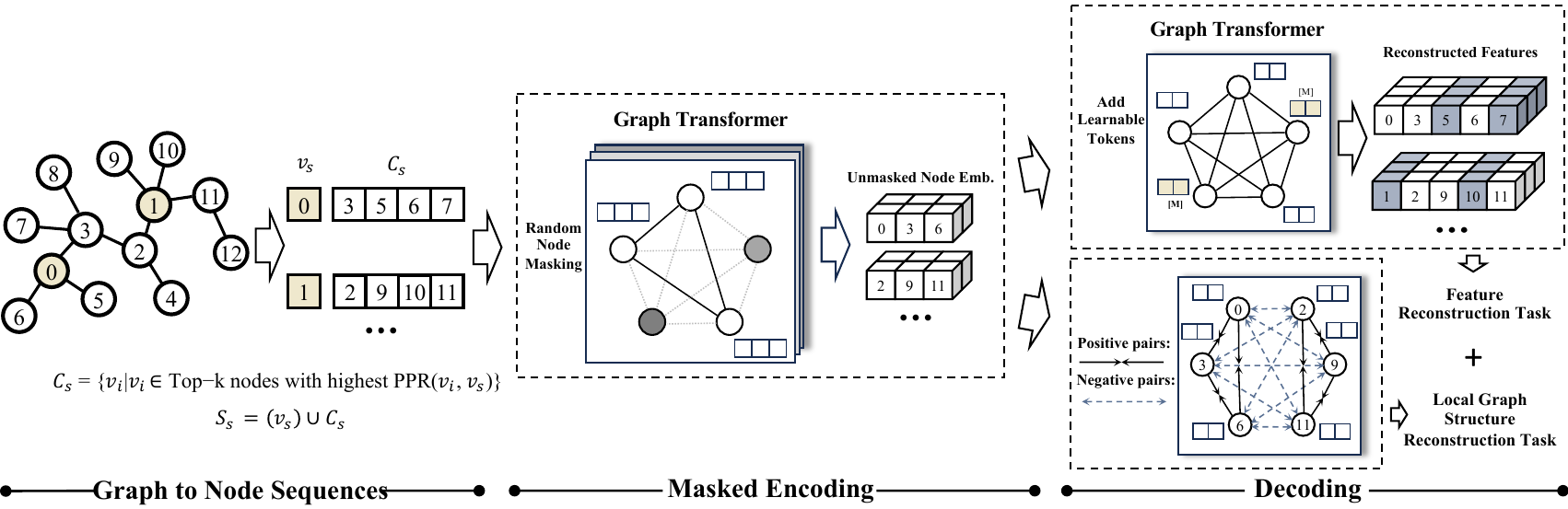}
    \vspace{-2mm}
    \caption{\textbf{Overview of \model framework during pre-training phase.} \textmd{(1) Given a seed node $v_s$, we adopt the personalized PageRank (PPR) algorithm to sample a contextual node sequence $S_s$ to represent its local graph structure. (2) Subsequently, we randomly mask a subset of nodes in each sequence and feed the sequence composed of the unmasked nodes into the graph transformer encoder. The output consists of the embeddings of each unmasked node. (3) The decoding comprises two learning objectives: a) we incorporate the masked nodes into the output of the encoder and initialize them with a learnable token $[M]$. A shallow graph transformer is employed as a decoder to reconstruct the input features of the masked nodes. b) In the context of graphs, each sequence $S_s$ can be interpreted as a local neighborhood or subgraph. We adopt a simple MLP and contrastive loss aiming to make nodes within each sequence similar in the latent space while making them dissimilar to nodes from other sequences. This encourages the learned embeddings to reflect the actual connectivity patterns within local graph structures.}
    }
    \label{fig:arch}
    \vspace{-5mm}
\end{figure*}
In this section, we present our proposed \model framework, which utilizes the MGM pretext task to pre-train graph transformers. The overall framework is illustrated in Fig.~\ref{fig:arch}.

\subsection{Graph to Node Sequences}
Due to the quadratic computational and memory complexity of self-attention with respect to the input sequence length, we are unable to scale it to large-scale graphs, such as social networks.
In this paper, we propose to sample contextual node sequences for each given seed node.
Specifically, we utilize the personalized PageRank (PPR) algorithm to sample nodes containing the most relevant information for each seed node as the sequence input for the Transformer model.
We define the following mathematical procedure: Given a graph represented as \( G = (V, E) \), we consider a set of seed nodes \( \boldsymbol{S} \) as input. The transition matrix \( \boldsymbol{P} \) is constructed as \( \boldsymbol{P}_{ij} = \frac{\boldsymbol{A}_{ij}}{\boldsymbol{d}_i} \), where \( \boldsymbol{A} \) is the adjacency matrix of \( G \), \( \boldsymbol{d}_i \) is the out-degree of node \( v_i \), and \( \boldsymbol{P}_{ij} \) represents the probability of transitioning from node \( v_i \) to node \( v_j \). Next, we calculate the personalized PageRank scores for each node in the graph by solving the equation \( \boldsymbol{r} = (1 - \alpha) \boldsymbol{r}_0 + \alpha \boldsymbol{P}^T \boldsymbol{r} \), where \( \boldsymbol{r} \) is the personalized PageRank vector, \( \boldsymbol{r}_0 \) is the initial personalized vector with \( \boldsymbol{r}_0(i) = 1 \) for \( i \) in \( \boldsymbol{S} \) and \( \boldsymbol{r}_0(i) = 0 \) otherwise, and \( \alpha \) is a damping factor.
After obtaining the personalized PageRank scores, we select contextual nodes for each seed node \( v_s \) based on the top-k nodes with the highest personalized PageRank scores: 
   \[
   C_s = \{v_{s,i} \,|\, v_{s,i} \in \text{Top-}k \text{ nodes with highest } \mathrm{PPR}(v_{s,i}, v_s)\}.
   \]
We define \( S_s \) as a node sequence, then \( S_s \) can be formulated as follows:
\(S_s = (v_s) \cup C_s.\)
Here, \( \cup \) denotes the operation of sequence concatenation. \( (v_s) \) is a singleton sequence containing just the source node \( v_s \), and \( C_s \) is a sequence of nodes \( (v_{s,1}, v_{s,2}, \ldots, v_{s,k}) \) such that \( C_s = \{v_{s,1}, v_{s,2}, \ldots, v_{s,k}\} \). 

\begin{figure}[t]
\centering
\includegraphics[width=\linewidth]{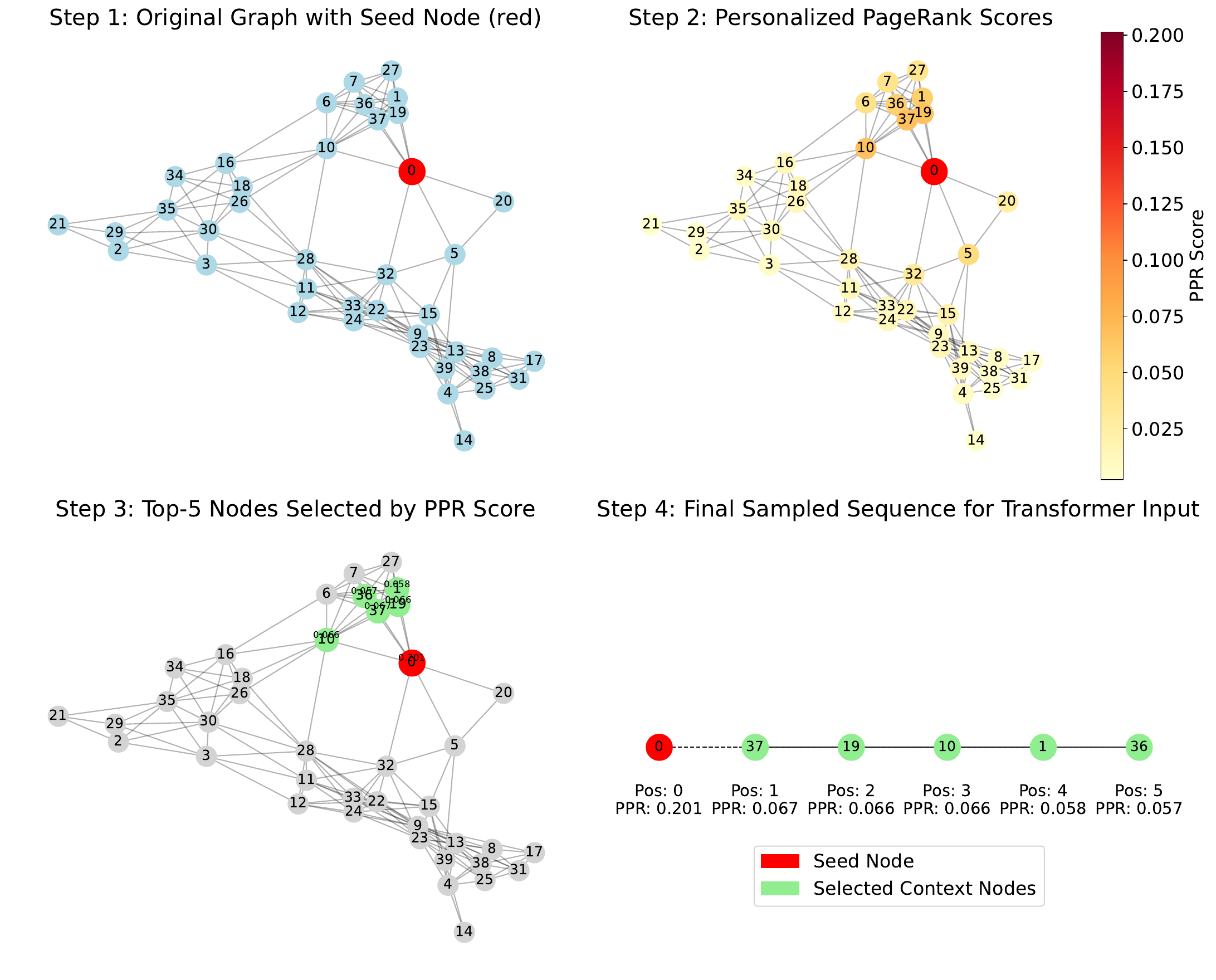}
\caption{\textbf{Step-by-step illustration of the PPR sampling process: }(1) Original graph with seed node, (2) Computation of Personalized PageRank scores, (3) Selection of top-$k$ nodes, and (4) Final sequence construction for transformer input.}
\vspace{-5mm}
\label{fig:ppr_sampling}
\end{figure}

\begin{figure}[t]
\centering
\includegraphics[width=\linewidth]{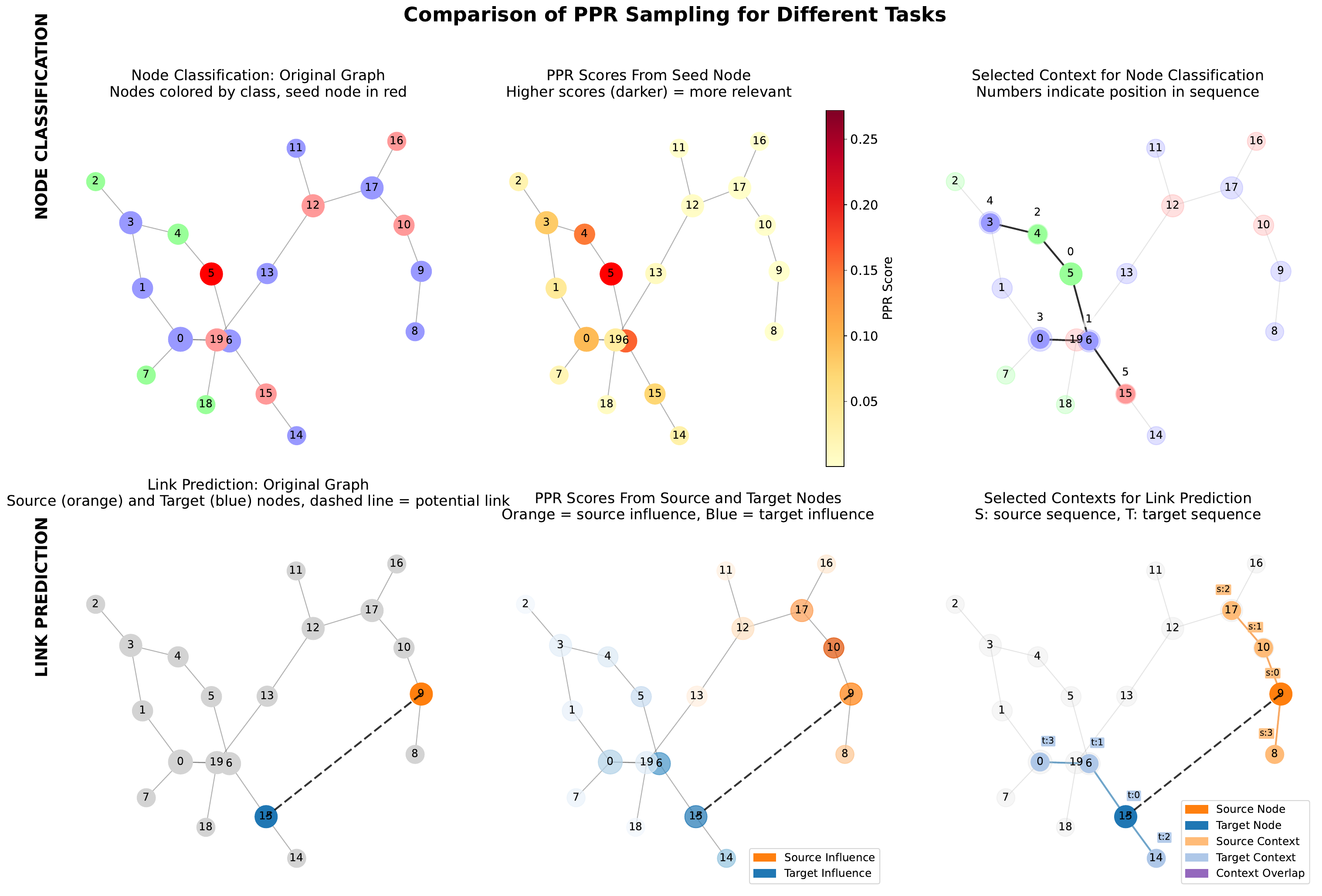}
\caption{\textbf{Comparison of PPR sampling for different tasks.} Top row: Node classification task with a single seed node. Bottom row: Link prediction task with source and target nodes, showing how contexts are sampled from both endpoints.}
\vspace{-5mm}
\label{fig:task_comparison}
\end{figure}

As illustrated in Fig.~\ref{fig:ppr_sampling}, PPR sampling follows a four-step process: (1) starting with a seed node, (2) computing importance scores that balance both proximity and centrality, (3) selecting the top-$k$ nodes with highest scores, and (4) constructing an ordered sequence for transformer input. Unlike traditional distance-based sampling approaches that may select unimportant neighbors or miss influential nodes slightly further away, PPR naturally identifies nodes that are both structurally important and meaningfully connected to the seed node. This creates balanced contextual sequences that effectively represent the local graph topology. As demonstrated in Fig.~\ref{fig:task_comparison}, this sampling approach is adaptable across different downstream tasks: for node classification (top row), PPR samples contexts that help determine class membership, while for link prediction (bottom row), we sample from both endpoints of the potential edge, with the overlap between contexts providing valuable signal for prediction. 

This sampling strategy offers several advantages: (1) It can be regarded as utilizing graph structural information to perform masking on full-graph attention, implicitly introducing inductive bias. (2) It unifies the sampling strategies for training and inference. Unlike sampling algorithms like GraphSAINT~\cite{zeng2019graphsaint} and GraphSAGE~\cite{hamilton2017inductive}, which mainly focus on training and employ full neighborhood aggregation during inference, our method enables the use of the same node sequences during both the training and the inference phases. (3) It decouples the sampling of labeled nodes and auxiliary nodes, as our method only needs to extract the corresponding sequences, and the time cost scales with the number of labeled nodes.

For pre-training, we randomly select 10\% of all nodes as seed nodes to ensure computational efficiency while maintaining representative coverage of the graph. During inference for downstream tasks, the seed nodes are simply the nodes requiring predictions, which effectively decouples the sampling of labeled nodes (typically scarce) from auxiliary nodes, enhancing efficiency in industrial applications.

\subsection{Masked Encoding}
Unlike most existing masked autoencoder-based graph self-supervised learning frameworks that employ GNNs as encoders~\cite{hou2022graphmae,li2022maskgae}, in this work, we use the Transformer model as the encoder, which introduces differences in the pipeline.

For a seed node \(v_s\) in the graph, we sample a sequence of contextual nodes $S_s = (v_s, v_{s,1}, v_{s,2}, \ldots, v_{s,k})$ of length $k+1$ including seed node \(v_s\) itself. A subset of this contextual sequence is randomly masked, resulting in a masked set $S^m_s = \{v_{s,m_1}, v_{s,m_2}, \ldots, v_{s,m_l}\}$ of length $l$, where \( l < k+1 \). The remaining unmasked nodes are represented as $S^u_s = \{v_{s,u_1}, v_{s,u_2}, \ldots, v_{s,u_{k-l+1}}\}$. 
The node feature $\mathbf{x}_{s,u_j} \in \mathbb{R}^{1 \times d}$ of each unmasked node \(v_{s,u_j}\) is first mapped into an embedding space using a linear projection from $d$ dimension to $d'$ dimension, where \(0 < j \leq k-l+1 \). 
In addition, we add pre-computed positional encodings to the node embedding:
\begin{equation}
    \mathbf{h}_{s,u_j} = \mathrm{Linear}(\mathbf{x}_{s,u_j}) + \mathbf{p}_{s,u_j}
\end{equation}
These embeddings serve as the input to the transformer. 
The embedding matrix of unmasked node sequence \( \mathbf{H}^{u}_{s} \in \mathbb{R}^{(k-l+1)\times d'} \)  are passed through a $L$-layer transformer encoder. In each layer, the transformer computes self-attention weights and aggregates information from all nodes in the sequence. The output after the final encoder layer is \( \mathbf{H}^{u}_{s} \), which represents the hidden representation of unmasked node sequence $S^u_s$ within the transformer. The specific operations in a transformer layer can be described as follows:$\mathbf{H}^{u}_{s} = f_E(\mathbf{H}^{u}_{s})$, where \( f_E(\cdot) \) represents the operations of the encoder of \model, which is a $L$-layers transformer encoder.

When using a GNN as an encoder, masked encoding still relies on the complete graph as input. However, when using a transformer as the encoder, we only need to encode a small portion of nodes (e.g., 15-25\%) that have not been masked, significantly improving training efficiency.

\subsection{Pre-training Tasks}
\vpara{Feature reconstruction.}
Given a node sequence $S_s$ along with its corresponding node features $\mathbf{X}_s \in \mathbb{R}^{(k+1) \times d}$, after the masked encoding phase, we obtain representations $\mathbf{H}^{u}_{s} \in \mathbb{R}^{(k-l+1)\times d'}$ for the unmasked node sequence. 
Similar to MIM~\cite{he2022masked}, the input to the decoder is the full set of tokens consisting of encoded unmasked nodes and masked nodes.
We employ a shared learnable token $\mathbf{h}_{[M]} \in \mathbb{R}^{1 \times d'}$ to initialize the representations of these masked nodes, and this learnable token is continually updated during training. At the same time, we once again add 
pre-computed positional encodings into the corresponding node embeddings to introduce graph structural information to the decoding phase.
The input representation matrix $\mathbf{H}_{s} \in \mathbb{R}^{(k+1)\times d'}$ of the decoder can be denoted as:
$$
    \mathbf{h}_{s,i} = 
    \begin{cases}
    \mathbf{h}_{[M]} + \mathbf{p}_{s,i} & v_{s,i} \in S^m_s \\
    \mathbf{h}^{u}_{s,i} + \mathbf{p}_{s,i} & v_{s,i} \notin S^m_s
    \end{cases}.
$$
Then, the decoder reconstructs the input features $\mathbf{X}_s$ from the latent embeddings $\mathbf{H}_{s}$. In this task, we use another relatively shallow transformer as a decoder, denoted as \( f_{D_{1}}(\cdot) \). The reconstructed features are $\mathbf{Z}_s = f_{D_{1}}(\mathbf{H}_{s})$. We adopt scaled cosine error~\cite{hou2022graphmae} as the objective function for feature reconstruction:
\begin{equation}
    \mathcal{L}_{1} = \frac{1}{|S^m_s| }\sum_{v_{s,i} \in S^m_s} (1 - \frac{\mathbf{x}_i^\top \mathbf{z}_i}{\lVert \mathbf{x}_i \lVert \cdot \lVert \mathbf{z}_i \lVert })^{\gamma}
\end{equation}
where $\mathbf{x}_i$ is the $i$-th row of input features $\mathbf{X}_s$, $\mathbf{z}_i$ is the $i$-th row of reconstructed feature $\mathbf{Z}_s$, and $\gamma \ge 1$ is the scaled coefficient. 

The feature reconstruction process in our framework operates similarly to filling in missing parts of an image based on visible portions. For each seed node, after identifying its contextual neighborhood and masking approximately 85\% of the nodes, the transformer encoder analyzes the few visible nodes to learn their relationships in the graph's latent space. The decoder then attempts to "imagine" what the hidden nodes' features should be, based solely on three sources of information: the processed embeddings from visible nodes, positional information about the masked nodes in the sequence, and a generic learnable token ([M]) indicating a hidden node. By comparing these predictions with the actual features and minimizing the difference, the model learns to understand feature relationships within local neighborhoods, recognize common feature patterns, and develop an implicit understanding of how structural positions correlate with node attributes. 

\vpara{Local graph structure reconstruction.}
Our intuition behind using PPR to sample contextual node sequences is that selecting the most relevant neighboring nodes can create small and densely connected clusters.
PPR can be interpreted within the spectral framework as a localized spectral method~\cite{green2021statistical}. 
Spectral clustering essentially operates in the eigenspace of the graph Laplacian matrix, capturing the low-frequency components that represent tightly-knit communities.~\cite{tremblay2020approximating}. 
Similarly, the embeddings learned by the proposed \model should ideally reside in a subspace where the local structural properties are maintained.
This task aims to capture local graph structure patterns by leveraging the likelihood that nodes within the same sequence exhibit higher similarity, in contrast to nodes from different sequences, which are generally less similar.

Defining \(N\) as the total number of sequences in a batch, 
\(M\) as the number of unmasked nodes in each sequence, 
\(\mathbf{h}_{i,j}\) as the representation vector of the \(j\)-th unmasked node in the \(i\)-th sequence after masked encoding. 
The sets of positive pairs \(\mathcal{P}\) and negative pairs \(\mathcal{N}\) are defined as follows:
\[
\begin{aligned}
  \mathcal{P} &= \{(\mathbf{h}_{i,j}, \mathbf{h}_{i,k}) : i \in \{1,...N\}, j, k \in \{1,...M_i\}, j \neq k \}, \\
  \mathcal{N} &= \{(\mathbf{h}_{i,j}, \mathbf{h}_{l,m}) : i, l \in \{1,...N\}, i \neq l,      \\
  &j \in \{1,...M_i\}, m \in \{1,...M_l\}\}.
\end{aligned}
\]
From \(\mathcal{P}\) and \(\mathcal{N}\), we stochastically sample subsets \(\tilde{\mathcal{P}} \subseteq \mathcal{P}\) and \(\tilde{\mathcal{N}} \subseteq \mathcal{N}\), each of size \(T\).
The similarity function \(\text{sim}: \mathbb{R}^{d'} \times \mathbb{R}^{d'} \rightarrow \mathbb{R}\) maps each pair of node embeddings to a scalar. 
In this task, we use a simple Multi-layer perception (MLP) as a decoder, denoted as \(f_{D_2}(\cdot)\). It transforms each node embedding before computing the similarity. Formally, the similarity scores \(s_{a,b}\) are given by:
\begin{equation}
    s_{a,b} = \text{sim}\left(f_{D_2}(\mathbf{h}_{a}), f_{D_2}(\mathbf{h}_{b})\right)
\end{equation}
The InfoNCE loss encourages the similarity scores for positive pairs to be high and for negative pairs to be low. The stochastic variant of the InfoNCE loss function, \(\mathcal{L}_{2}\), is defined as:
\begin{equation}
\small
    \begin{aligned}
\mathcal{L}_{2} = -\frac{1}{T} \sum_{(\mathbf{h}_{i,j}, \mathbf{h}_{i,k}) \in \tilde{\mathcal{P}}} \log \left( \frac{\exp(s_{ij,ik})}{\sum_{(\mathbf{h}_{l,m}, \mathbf{h}_{n,o}) \in \tilde{\mathcal{P}} \cup \tilde{\mathcal{N}}} \exp(s_{lm,no})} \right).
\end{aligned}
\end{equation}

\vpara{Training.}
In this part, we summarize the overall training flow of \model. 
Given a web-scale graph, we first use the PPR algorithm to sample a contextual node sequence for each seed node. 
For pre-training, for efficiency reasons, we generally only need to select a subset of all nodes to sample their corresponding sequences as pre-training data. 
After being randomly masked, these sequences are then passed through the Transformer-based encoder to generate latent representations of the unmasked nodes.
Next, we use the two different decoders to reconstruct the original graph information from them. 
Then we obtain the overall loss by fusing the two losses with a mixing coefficient $\lambda$:
\begin{equation}
    \mathcal{L} = \mathcal{L}_{1} + \lambda \cdot \mathcal{L}_{2}
\end{equation}

\subsection{Decoder Reuse for Feature Augmentation}
\begin{figure}
    \centering
    \includegraphics[width=0.9\linewidth]{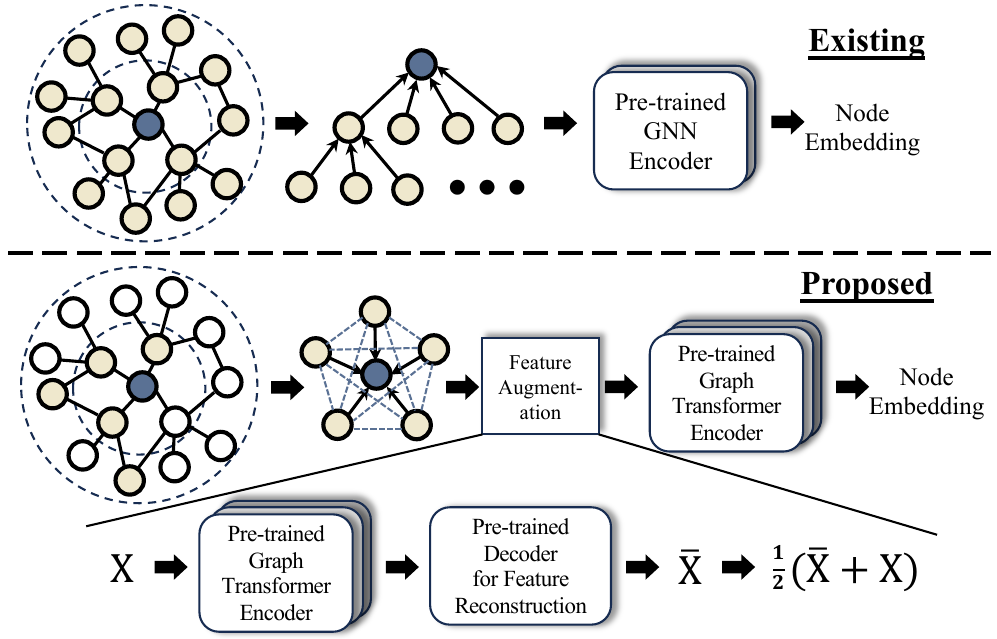}
    \caption{\textbf{Comparison of existing methods and the proposed \model framework during the inference phase.} \textmd{Our contributions are: a) we use fast approximations of PPR to select informative auxiliary nodes, instead of employing layer-by-layer full neighborhood aggregation. b) We propose a lightweight feature augmentation strategy that requires no additional training. Using the pre-trained encoder and decoder to perform a single forward pass generates reconstructed features, which, when averaged with the original features, serve as the input for downstream tasks.}}
    \label{fig:infer}
    \vspace{-5mm}
\end{figure}
Under the classic auto-encoder paradigm, only the pre-trained encoder is used in downstream tasks, and the decoder is discarded~\cite{he2022masked,hou2022graphmae,hou2023graphmae2,li2022maskgae}. 
In \model, we propose a novel approach to utilize the pre-trained decoder for feature augmentation when the pre-trained encoder is used for inference or finetuned for downstream tasks.
Given a contextual node sequence $S_q$ of a query node $v_q$,
we generate reconstructed node features by passing the node sequences and their features $\mathbf{X'_q} \in \mathbb{R}^{(k+1) \times d}$ into the pre-trained encoder and decoder for features reconstruction without any mask: 
\begin{equation}
    \hat{\mathbf{H'_q}} = f_E( \mathrm{Linear}(\mathbf{X'_q}) + \mathbf{P'_q})
\end{equation}
\begin{equation}
    \overline{\mathbf{X'_q}} = f_{D_{1}}(\hat{\mathbf{H'_q}} + \mathbf{P'_q})
\end{equation}
\noindent where $\mathbf{P'_q} \in \mathbb{R}^{(k+1) \times d'}$ is the pre-computed positional encodings matrix and $\overline{\mathbf{X'_q}} \in \mathbb{R}^{(k+1) \times d}$ is the reconstructed features matrix. 
Then we average $\mathbf{X'_q}$ and $\overline{\mathbf{X'_q}}$ to serve as input for the pre-trained encoder during fine-tuning or inference. 

This design is motivated by our argument that, despite the success of GNNs and Graph Transformers in learning node representations from local neighborhoods, the limited number of neighbors can constrain their performance.
Our idea is to utilize the node feature patterns learned by masked auto-encoders to generate additional node features to augment the original ones.

\vpara{Inference.}
Fig.~\ref{fig:infer} shows the comparison between existing methods and the proposed \model framework during the inference phase.
The process of utilizing pre-trained decoder for feature augmentation is decoupled from the subsequent fine-tuning of encoder for downstream tasks. After obtaining the newly augmented features, decoders are discarded, and only the encoder is employed for generating node representations or subject to fine-tuning. We take the output of the last layer of encoder as node embeddings:
\begin{equation}
    \mathbf{H'_q} = f_E( \mathrm{Linear}(\frac{1}{2}(\mathbf{X'_q} + \overline{\mathbf{X'_q}})) + \mathbf{P'_q})
\end{equation}

When fine-tuning the model for specific downstream tasks, classification heads are added after the encoder according to the task requirements.
In prior research endeavors, most sampling-based scalable graph neural networks have solely focused on training efficiency. 
In inference, they aggregate information from all neighbors for each node, utilizing full graph information, which leads to low efficiency.
In this work, we bridge the efficiency gap between training and inference. By adopting the PPR algorithm to sample contextual node sequences in both the training and the inference phases, seed nodes can aggregate information from the select few neighbors with the richest information.

%% file: 4.experiments.tex
\section{Performance on Public Benchmarks}
\model is designed as a general graph pre-training framework. Before deploying our method on Tencent, we conduct experiments on public benchmarks to validate its effectiveness and broad applicability.
\subsection{Pre-training}
\begin{table}[t]
\centering
\caption{\label{tab:public_dataset}Statistics of public datasets.}
\renewcommand\tabcolsep{3.2pt}
\begin{tabular}{l|rrrr}
\toprule[1.1pt]
Datasets & \#Nodes  & \#Edges & \#Features & Domain \\
\midrule
Cora     & 2,708  &  5,429 & 300  & Citation \\
PubMed   & 19,717  & 44,338 & 300  & Citation \\
ogbn-arxiv       & 169,343 & 1,166,243 & 300 & Citation \\
ogbn-papers100M  & 111,059,956 & 1,615,685,872 & 300 & Citation \\
ogbn-products & 2,449,029 & 61,859,140 & 300 & E-commerce \\
Wiki-CS & 11,701 & 216,123 & 300 & Web content \\
FB15K237 & 14,541 & 310,116 & 300 & Knowledge \\
WN18RR & 40,943 & 93,003 & 300 & Knowledge \\
\bottomrule[1.1pt]
\end{tabular}
\vspace{-4mm}
\end{table}

\vpara{Dataset.}
We conduct pre-training on a large paper citation network ogbn-papers100M~\cite{hu2021ogb}.
In this dataset, each node represents a paper and each directed edge indicates a citation. Each node is associated with its natural language title and abstracts. 
We leverage word2vec~\cite{mikolov2013distributed} to generate unified vector features for all nodes.
The statistics are listed in Table~\ref{tab:public_dataset}.

\vpara{Baselines.}
The baselines are consistent with those on Tencent's data in Section~\ref{sec:base}.
To maximize the performance of the baselines, we adopt the widely used GraphSAINT~\cite{zeng2019graphsaint} as the subgraph sampling method in our baselines. 
For all baseline methods, we employ GAT~\cite{velivckovic2017graph} as the backbone network.

\subsection{In-domain Node Classification Task}

\begin{table}[t]
\centering
\caption{\label{tab:pnc} Linear probing results on Node Classification Task. \textmd{We report accuracy(\%) for all datasets.
No-pretrain represents a random-initialized model without any pre-training.}}
\begin{tabular}{l|ccc}
\toprule[1.1pt]
Methods &  Cora  & PubMed & ogbn-arxiv \\
\midrule 
MLP & 55.34$\pm$1.31 & 63.48$\pm$0.54 & 55.50$\pm$0.23 \\
\midrule 
No-pretrain$_{\small G\hspace{-0.5mm}A\hspace{-0.5mm}T}$ & 64.42$\pm$0.45 & 66.18$\pm$1.24 & 67.12$\pm$0.27 \\
DGI & 61.24$\pm$0.24 & 63.29$\pm$1.46 & 64.11$\pm$0.22 \\
GRACE & 63.24$\pm$1.24 & 64.18$\pm$1.22 & \underline{69.29$\pm$0.22} \\
BGRL & 66.41$\pm$0.76 & 63.24$\pm$1.45 & 68.24$\pm$0.52 \\
GraphMAE & 67.52$\pm$0.34 & 66.15$\pm$1.53 & 68.12$\pm$0.29 \\
GraphMAE2 & \underline{67.97$\pm$0.56} & \underline{68.98$\pm$1.42} & 68.01$\pm$1.22 \\
\midrule
No-pretrain$_{\small T\hspace{-0.4mm}r\hspace{-0.4mm}a\hspace{-0.4mm}n\hspace{-0.4mm}s\hspace{-0.4mm}f\hspace{-0.4mm}o\hspace{-0.4mm}r\hspace{-0.4mm}m\hspace{-0.4mm}e\hspace{-0.4mm}r}$ & 65.01$\pm$1.22 & 65.83$\pm$1.01 & 67.45$\pm$0.43 \\
\model & \textbf{70.11$\pm$0.44} & \textbf{69.24$\pm$1.21} & \textbf{70.22$\pm$0.42} \\
\scriptsize Improv. vs Best Baseline (\%) & 3.14\% & 0.38\% & 1.35\% \\
\bottomrule[1.1pt]
\end{tabular}
\vspace{-5mm}
\end{table}

\vpara{Setup.}
In this study, we adopt Cora~\cite{mccallum2000automating}, PubMed~\cite{sen2008collective} and ogbn-arxiv~\cite{hu2021ogb}, three popular citation network benchmarks for node classification. 
In these datasets, each node represents a paper and each directed edge indicates a citation. Each node is associated with its natural language title and abstracts. 
In alignment with the pre-training graph, we leverage word2vec~\cite{mikolov2013distributed} to generate unified vector features for all datasets. The statistics are listed in Table~\ref{tab:public_dataset}.
For Cora and PubMed, we follow the public data splits as in~\cite{velivckovic2018deep,hou2022graphmae,he2023explanations}.
For ogbn-arxiv, we follow the official splits in~\cite{hu2021ogb}.
We select all the self-supervised pre-trained models mentioned earlier as baseline methods. 
To maximize the performance of the baselines, we employ full neighborhood aggregation in inference.
We evaluate our method under \textit{linear probing} settings.

\vpara{Results.}
Table~\ref{tab:pnc} shows the results of our method and baselines under \textit{linear probing} setting. We can observe that our proposed \model framework generally outperforms all baselines across all three datasets. This substantiates that \model is a general and effective graph pre-training framework, capable of applicability across a diverse range of graph domains.
We also observe that some baseline methods manifest severe cases of \textit{negative transfer}.
We hypothesize that a potential reason could be that the citation networks themselves do not possess numerical features in the same feature space. The process of encoding textual features using pre-trained language models may introduce unavoidable noise. 
The noise in node features adversely impacts the generalization capability of the pre-trained models.
Our proposed method for feature augmentation, which utilizes reconstructed features based on masked autoencoders, can be regarded as a form of feature denoising~\cite{vincent2008extracting}. 
During inference, the model applies its learned error-correction ability to the unmasked input, benefiting from the robustness learned during the training phase~\cite{vincent2010stacked}.

\subsection{Cross-domain Evaluation on Diverse Tasks}
\begin{table}[t]
\renewcommand\tabcolsep{2.8pt}
\centering
\caption{Linear probing results on datasets from diverse domains. We report accuracy (\%) for both node classification tasks (ogbn-products, Wiki-CS) and edge classification tasks (FB15K237, WN18RR). No-pretrain represents a random-initialized model without any pre-training.}
\label{tab:diverse_domains}
\begin{tabular}{lcccc}
\toprule[1.1pt]
\multirow{2}{*}{Methods} & \multicolumn{2}{c}{Node Classification} & \multicolumn{2}{c}{Edge Classification} \\
\cmidrule(lr){2-3} \cmidrule(lr){4-5}
& ogbn-products & Wiki-CS & FB15K237 & WN18RR \\
\midrule
MLP & 66.29$\pm$0.21 & 66.23$\pm$0.11 & 66.11$\pm$0.45 & 63.03$\pm$0.54 \\
\midrule
No-pretrain$_{\small G\hspace{-0.5mm}A\hspace{-0.5mm}T}$ & 67.11$\pm$0.34 & 70.11$\pm$0.67 & 66.98$\pm$0.26 & 62.80$\pm$0.64 \\
DGI & 64.21$\pm$0.32 & 67.11$\pm$0.12 & 26.99$\pm$0.22 & 52.04$\pm$0.22 \\
GRACE & 66.85$\pm$0.28 & 68.54$\pm$0.29 & 60.21$\pm$0.25 & 58.32$\pm$0.32 \\
BGRL & 66.22$\pm$0.39 & 70.12$\pm$0.15 & 64.91$\pm$0.22 & 56.44$\pm$0.21 \\
GraphMAE & 67.19$\pm$0.39 & 68.11$\pm$0.12 & 61.11$\pm$0.12 & 59.76$\pm$0.29 \\
GraphMAE2 & 67.73$\pm$0.12 & 68.84$\pm$0.37 & 63.76$\pm$0.12 & 60.24$\pm$0.23 \\
\midrule
No-pretrain$_{\small T\hspace{-0.4mm}r\hspace{-0.4mm}a\hspace{-0.4mm}n\hspace{-0.4mm}s.}$ & 67.98$\pm$0.34 & 69.14$\pm$0.29 & 65.31$\pm$0.25 & 61.42$\pm$0.31 \\
PGT & \textbf{70.53$\pm$0.23} & \textbf{72.32$\pm$0.22} & \textbf{68.47$\pm$0.28} & \textbf{64.81$\pm$0.26} \\
\midrule
\scriptsize Improv. (\%) & 4.13\% & 5.05\% & 5.65\% & 7.59\% \\
\bottomrule[1.1pt]
\end{tabular}
\vspace{-5mm}
\end{table}

\begin{table}[t]
\centering
\caption{Dynamic link prediction results on the time-split \textbf{ogbn-arxiv} dataset. We report MRR (\%) and Hits@k (\%). Static baselines are trained on the aggregated historical graph.}
\label{tab:dynamic_lp}
\begin{tabular}{lccc}
\toprule[1.1pt]
Method & MRR & Hits@10 & Hits@50 \\
\midrule
\textit{Static Baselines} \\
GCN~\cite{kipf2016semi}         & 32.14 & 46.81 & 66.53 \\
GraphSAGE~\cite{hamilton2017inductive} & 33.59 & 48.72 & 68.21 \\
GAT~\cite{velivckovic2017graph}         & 34.72 & 50.15 & 70.09 \\
\midrule
\textit{Dynamic Baselines} \\
EvolveGCN~\cite{pareja2020evolvegcn} & 35.81 & 51.34 & 71.82 \\
DGCN~\cite{gao2022novel}      & 36.54 & 52.01 & 72.45 \\
VGRNN~\cite{hajiramezanali2019variational}     & 37.92 & 54.66 & 74.13 \\
TGN~\cite{rossi2020temporal}        & \underline{41.88} & \underline{60.29} & \underline{80.54} \\
\midrule
\textbf{PGT-Dynamic (Ours)} & \textbf{45.12} & \textbf{64.77} & \textbf{84.36} \\
\midrule
\scriptsize Improv. vs Best Baseline (\%) & 7.74\% & 7.43\% & 4.74\% \\
\bottomrule[1.1pt]
\end{tabular}
\vspace{-3mm}
\end{table}
\begin{table}[t]
\centering
\caption{\label{tab:pdata}Statistics of datasets for pre-training and Minor Detection Task.}
\begin{tabular}{l|rrc}
\toprule[1.1pt]
Datasets & \#Nodes  & \#Edges & \#Features\\
\midrule 
pretraining-graph  & 547,358,876  &  12,243,334,598 & 146  \\
Game A & 112,034,422 & 1,824,946,632 & 146 \\ 
Game B & 32,234,396 & 763,463,297 & 146 \\ 
\bottomrule[1.1pt]
\end{tabular}
\vspace{-4mm}
\end{table}

\begin{table}[t]
\centering
\caption{\label{tab:lnc} Linear probing results on Minors Detection Task. \textmd{We report ROC-AUC for all datasets.
No-pretrain represents a random-initialized model without any pre-training.}}
\begin{tabular}{l|cc}
\toprule[1.1pt]
Methods &  Game A  & Game B  \\
\midrule 
MLP & 0.6461 $\pm$ 0.0012 & 0.7239 $\pm$ 0.0013 \\
\midrule 
No-pretrain$_{\small G\hspace{-0.5mm}A\hspace{-0.5mm}T}$ & 0.6621 $\pm$ 0.0013 & 0.7829 $\pm$ 0.0003 \\
DGI & 0.6599 $\pm$ 0.0032 & 0.7123 $\pm$ 0.0023 \\
GRACE & 0.6745 $\pm$ 0.0013 & 0.7739 $\pm$ 0.0011 \\
BGRL & 0.6681 $\pm$ 0.0015 & 0.7789 $\pm$ 0.0011 \\
GraphMAE & \underline{0.6781 $\pm$ 0.0012} & 0.7827 $\pm$ 0.0009 \\
GraphMAE2 & 0.6765 $\pm$ 0.0011 & \underline{0.7987 $\pm$ 0.0006} \\
\midrule
No-pretrain$_{\small T\hspace{-0.4mm}r\hspace{-0.4mm}a\hspace{-0.4mm}n\hspace{-0.4mm}s\hspace{-0.4mm}f\hspace{-0.4mm}o\hspace{-0.4mm}r\hspace{-0.4mm}m\hspace{-0.4mm}e\hspace{-0.4mm}r}$ & 0.6771 $\pm$ 0.0021 & 0.8021 $\pm$ 0.0001 \\
\model & \textbf{0.7087 $\pm$ 0.0011} & \textbf{0.8121 $\pm$ 0.0013} \\
\scriptsize Improv. vs Best Baseline (\%) & 4.50\% & 1.24\% \\
\bottomrule[1.1pt]
\end{tabular}
\vspace{-5mm}
\end{table}

\begin{table*}
\vspace{-7mm}
    \centering
    \caption{Results of fine-tuning the pre-trained model with 1\%, 5\% and 100\% labeled training data on Minors Detection Task. \textmd{We report ROC-AUC for all datasets. No-pretrain represents a random-initialized model without any pre-training.}}
    \vspace{-2mm}
    \renewcommand\tabcolsep{4.5pt}
    \scalebox{0.8}{
    \begin{tabular}{l|ccc|ccc}
    \toprule[1.1pt]
                & \multicolumn{3}{c|}{Game A} & \multicolumn{3}{c}{Game B} \\
    \midrule
    Label Ratio & 1\% &5\% &100\% & 1\% &5\% &100\%\\
    \midrule
    No-pretrain$_{\small G\hspace{-0.5mm}A\hspace{-0.5mm}T}$ & 0.6441 $\pm$ 0.0329 & 0.6583 $\pm$ 0.0274 & 0.7031 $\pm$ 0.0023 & 0.7221 $\pm$ 0.0214 & 0.7623 $\pm$ 0.0212 & 0.8101 $\pm$ 0.0021 \\
    DGI & 0.6632 $\pm$ 0.0212 & 0.6721 $\pm$ 0.0201 & 0.6986 $\pm$ 0.0043 & 0.7231 $\pm$ 0.0345 & 0.7732 $\pm$ 0.0112 & 0.8009 $\pm$ 0.0031 \\
    GRACE & 0.6587 $\pm$ 0.0197 & 0.6723 $\pm$ 0.0097 & 0.7056 $\pm$ 0.0021 & 0.7348 $\pm$ 0.0214 & 0.7786 $\pm$ 0.0085 & 0.8097 $\pm$ 0.0018 \\
    BGRL & 0.6701 $\pm$ 0.0101 & 0.6792 $\pm$ 0.0119 & 0.7132 $\pm$ 0.0045 & 0.7438 $\pm$ 0.0100 & 0.7829 $\pm$ 0.0097 & 0.8187 $\pm$ 0.0022 \\
    GraphMAE & \textbf{0.6772 $\pm$ 0.0132} & \underline{0.6842 $\pm$ 0.0103} & 0.7128 $\pm$ 0.0024 & 0.7419 $\pm$ 0.0098 & 0.7855 $\pm$ 0.0087 & 0.8176 $\pm$ 0.0032 \\
    GraphMAE2 & 0.6746 $\pm$ 0.0119 & 0.6821 $\pm$ 0.0056 & 0.7132 $\pm$ 0.0034 & \underline{0.7450 $\pm$ 0.0129} & \underline{0.7932 $\pm$ 0.0035} & 0.8211 $\pm$ 0.0024 \\
    \midrule
    No-pretrain$_{\small T\hspace{-0.4mm}r\hspace{-0.4mm}a\hspace{-0.4mm}n\hspace{-0.4mm}s\hspace{-0.4mm}f\hspace{-0.4mm}o\hspace{-0.4mm}r\hspace{-0.4mm}m\hspace{-0.4mm}e\hspace{-0.4mm}r}$ & 0.6349 $\pm$ 0.0437 & 0.6538 $\pm$ 0.0357 & \underline{0.7193 $\pm$ 0.0023} & 0.7238 $\pm$ 0.0322 & 0.7863 $\pm$ 0.0129 & \underline{0.8222 $\pm$ 0.0034} \\
    \model & \underline{0.6753 $\pm$ 0.0103} & \textbf{0.6887 $\pm$ 0.0126} & \textbf{0.7301 $\pm$ 0.0021} & \textbf{0.7532 $\pm$ 0.0213} & \textbf{0.8083 $\pm$ 0.0187} & \textbf{0.8329 $\pm$ 0.0076} \\
    \scriptsize Improv. vs Best Baseline (\%) & -0.28\% & 0.66\% & 1.50\% & 1.10\% & 1.90\% & 1.30\% \\
    \bottomrule[1.1pt]
    \end{tabular}
    }
    \label{tab:fnc}
    \vspace{-3mm}
\end{table*}

\vpara{Setup.} To evaluate the broad applicability of PGT beyond citation networks and gaming scenarios, we conduct additional experiments on four datasets from diverse domains: ogbn-products, Wiki-CS, FB15K237, and WN18RR. These datasets represent varied application scenarios in e-commerce, web content networks, and knowledge graphs. For ogbn-products, each node represents a product sold on Amazon, and edges indicate that the products are frequently purchased together. Wiki-CS contains Wikipedia articles about computer science topics, with edges representing hyperlinks between articles. FB15K237 and WN18RR are knowledge graph datasets where nodes represent entities and edges represent relationships between them. Following our main experimental protocol, we pre-train on ogbn-papers100M and evaluate performance on these target datasets using linear probing. For node classification tasks (ogbn-products and Wiki-CS), we train a linear classifier on top of the frozen node embeddings. For edge classification tasks (FB15K237 and WN18RR), we follow standard practice and concatenate the embeddings of node pairs to train a linear classifier.
    
\vpara{Results.} Table \ref{tab:diverse_domains} shows the results on these diverse datasets. We observe that PGT consistently outperforms all baseline methods across different domains and tasks, with improvements ranging from 4.13\% to 7.59\% over the best baseline. This demonstrates the strong generalization capability of our pre-training framework beyond citation networks. Notably, on the knowledge graph datasets (FB15K237 and WN18RR), some baseline methods like DGI show significant performance degradation, likely due to the structural differences between the pre-training and target graphs. In contrast, PGT maintains robust performance, suggesting that our approach effectively mitigates negative transfer issues. The superior performance on e-commerce networks (ogbn-products) and web content networks (Wiki-CS) further confirms that PGT's pre-training strategy captures transferable knowledge applicable to diverse graph structures and domains. These results substantiate our claim that PGT is a general graph pre-training framework with broad applicability across various real-world scenarios beyond our primary application in the gaming industry.

\subsection{Dynamic Link Prediction Task}

To demonstrate the versatility of our framework and its applicability to temporal tasks, we adapt PGT for dynamic link prediction. We introduce \textbf{PGT-Dynamic}, a straightforward yet effective extension of our model. For each graph snapshot at time $t$, we first use the powerful, pre-trained PGT encoder (frozen) to generate high-quality node embeddings $H_t$ that capture the intra-snapshot structure. These embeddings are then fed into a Gated Recurrent Unit (GRU)~\cite{chung2014empirical}, which maintains a hidden state to aggregate temporal information across snapshots. The final, temporally-aware node representation at time $t$ is the output of the GRU. This hybrid approach leverages the rich features from our static pre-training while a lightweight recurrent model captures the temporal dynamics.

\vpara{Setup.}
We adapt the \textbf{ogbn-arxiv} dataset for a dynamic link prediction task. Using the publication year of each paper, we create yearly graph snapshots. The task is to predict new citations (links) in a future year based on the graph's evolution. We use snapshots from before 2017 for training, the 2018 snapshot for validation, and the 2019 snapshot for testing. We compare PGT-Dynamic against both static and dynamic GNN baselines. The static baselines (GCN~\cite{kipf2016semi}, GraphSAGE~\cite{hamilton2017inductive}, GAT~\cite{velivckovic2017graph}) are trained on an aggregated graph of all training snapshots. The dynamic baselines include EvolveGCN~\cite{pareja2020evolvegcn}, DGCN~\cite{gao2022novel}, TGN~\cite{rossi2020temporal}, and VGRNN~\cite{hajiramezanali2019variational}. These are trained sequentially on the snapshots. We evaluate all methods using standard link prediction metrics: Mean Reciprocal Rank (MRR) and Hits@k.

\vpara{Results.}
As shown in Table~\ref{tab:dynamic_lp}, our proposed PGT-Dynamic significantly outperforms all baselines. We observe that the standard static GNNs perform poorly, as they are incapable of capturing the temporal evolution of the graph, confirming the necessity of temporal modeling. Among the specialized dynamic methods, TGN is the strongest baseline. Even so, PGT-Dynamic surpasses TGN by a considerable margin (e.g., a 7.74\% relative improvement in MRR).

This result is highly compelling because the dynamic baselines are specifically designed to model temporal evolution from scratch, whereas our adaptation is a simple combination of a static encoder and a GRU. The superior performance of PGT-Dynamic stems from the powerful node representations generated by the pre-trained PGT encoder. These features, learned from the massive ogbn-papers100M dataset, provide a much richer semantic and structural starting point than the randomly initialized features used by the other models. The GRU can then focus on learning the temporal evolution from a high-quality representation space. This experiment strongly suggests that a powerful pre-trained static model can serve as a robust foundation for building effective dynamic models, highlighting the generalization and flexibility of the PGT framework.

\section{Deployment in Tencent}
We now introduce the implementation and deployment of the \model framework in Tencent, the largest online gaming company in China as well as in the world by revenues~\footnote{\url{https://newzoo.com/resources/rankings/top-25-companies-game-revenues}}.
\subsection{Pre-training}

\begin{table}[t]
\centering
\caption{\label{tab:fdata}Statistics of datasets for Friend Recall Task.}
\begin{tabular}{l|rrcc}
\toprule[1.1pt]
Datasets & \#Edges & \#Nodes & \#Features & \#Snapshots\\
\midrule 
Game C & 139,483,723 & 23,419,874 & 146 & 5\\ 
Game D & 232,454,298 & 16,732,498 & 146 & 5\\ 
\bottomrule[1.1pt]
\end{tabular}
\vspace{-5mm}
\end{table}

\begin{table*}
\vspace{-7mm}
    \centering
    \caption{Linear probing results on Friend Recall Task. \textmd{No-pretrain represents a random-initialized model without any pre-training.}}
    \renewcommand\tabcolsep{6.5pt}
    \scalebox{0.8}{
    \begin{tabular}{l|ccccc|ccccc}
    \toprule[1.1pt]
                & \multicolumn{5}{c|}{Game C} & \multicolumn{5}{c}{Game D} \\
    \midrule
    Metrics & Hits@1 & Hits@3 &Hits@5 & Hits@10 & MRR &Hits@1 & Hits@3 &Hits@5 & Hits@10 & MRR \\
    \midrule
    XGBoost & 0.1384 & 0.3645 & 0.4983 & 0.6726 & 0.2721 & 0.4421 & 0.6892 & 0.8032 & 0.9121 & 0.5922\\
    MLP & 0.1453 & 0.3784 & 0.5242 & 0.6921 & 0.2764 & 0.4529 & 0.6991 & 0.8039 & 0.9093 & 0.5998 \\
    Bilinear & 0.1423 & 0.3773 & 0.5212 & 0.6908 & 0.2734 & 0.4435 & 0.6998 & 0.8121 & 0.9123 & 0.5894\\
    \midrule
    No-pretrain$_{\small G\hspace{-0.5mm}A\hspace{-0.5mm}T}$ & 0.2343 & 0.4239 & 0.5912 & 0.7198 & 0.3211 & 0.4221 & 0.6732 & 0.7729 & 0.8621 & 0.5321\\
    DGI & 0.2512 & 0.4387 & 0.5894 & 0.7124 & 0.3267 & 0.4326 & 0.6784 & 0.7842 & 0.8731 & 0.5401\\
    GRACE & 0.2833 & 0.4721 & 0.6091 & 0.7532 & 0.3721 & 0.4601 & 0.7001 & 0.8175 & 0.9183 & 0.6034\\
    BGRL & \underline{0.2982} & \underline{0.4892} & \underline{0.6192} & \underline{0.7622} & 0.3894 & 0.4627 & 0.7031 & 0.8210 & 0.9234 & 0.6073\\
    GraphMAE & 0.2845 & 0.4724 & 0.6129 & 0.7601 & \underline{0.3921} & 0.4519 & 0.6921 & 0.8129 & 0.9139 & 0.5983 \\
    GraphMAE2 & 0.2783 & 0.4807 & 0.6091 & 0.7587 & 0.3899 & \underline{0.4729} & \underline{0.7082} & \underline{0.8215} & \underline{0.9257} & \underline{0.6164}\\
    \midrule
    No-pretrain$_{\small T\hspace{-0.4mm}r\hspace{-0.4mm}a\hspace{-0.4mm}n\hspace{-0.4mm}s\hspace{-0.4mm}f\hspace{-0.4mm}o\hspace{-0.4mm}r\hspace{-0.4mm}m\hspace{-0.4mm}e\hspace{-0.4mm}r}$ & 0.2454 & 0.4439 & 0.5991 & 0.7342 & 0.3398 & 0.4329 & 0.6793 & 0.7832 & 0.8821 & 0.5429\\
    \model & \textbf{0.3125} & \textbf{0.5023} & \textbf{0.6389} & \textbf{0.7942} & \textbf{0.4034} & \textbf{0.4894} & \textbf{0.7121} & \textbf{0.8298} & \textbf{0.9278} & \textbf{0.6273}\\
    \scriptsize Improv. vs Best Baseline (\%) & 4.80\% & 2.68\% & 3.18\% & 4.20\% & 2.88\% & 3.48\% & 0.55\% & 1.01\% & 0.23\% & 1.77\% \\
    \bottomrule[1.1pt]
    \end{tabular}
    }
    \label{tab:llp}
    \vspace{-4mm}
\end{table*}

\begin{table*}
    \centering 
    \caption{Results of fine-tuning the pre-trained model on Friend Recall Task. \textmd{No-pretrain represents a random-initialized model without any pre-training.}}
    \renewcommand\tabcolsep{6.5pt}
    \scalebox{0.8}{
    \begin{tabular}{l|ccccc|ccccc}
    \toprule[1.1pt]
                & \multicolumn{5}{c|}{Game C} & \multicolumn{5}{c}{Game D} \\
    \midrule
    Metrics & Hits@1 & Hits@3 &Hits@5 & Hits@10 & MRR &Hits@1 & Hits@3 &Hits@5 & Hits@10 & MRR \\
    \midrule
    XGBoost & 0.1384 & 0.3645 & 0.4983 & 0.6726 & 0.2721 & 0.4421 & 0.6892 & 0.8032 & 0.9121 & 0.5922\\
    MLP & 0.1453 & 0.3784 & 0.5242 & 0.6921 & 0.2764 & 0.4529 & 0.6991 & 0.8039 & 0.9093 & 0.5998 \\
    Bilinear & 0.1423 & 0.3773 & 0.5212 & 0.6908 & 0.2734 & 0.4435 & 0.6998 & 0.8121 & 0.9123 & 0.5894\\
    \midrule
    No-pretrain$_{\small G\hspace{-0.5mm}A\hspace{-0.5mm}T}$& 0.3245 & 0.5012 & 0.6382 & 0.8021 & 0.4102 & 0.4698 & 0.7028 & 0.8203 & 0.9213 & 0.6046 \\
    DGI & 0.3024 & 0.4821 & 0.6187 & 0.7920 & 0.4012 & 0.4424 & 0.6824 & 0.7934 & 0.8901 & 0.5587 \\
    GRACE & 0.3394 & 0.5329 & 0.6476 & 0.8102 & 0.4123 & 0.4712 & 0.7132 & 0.8245 & 0.9253 & 0.6088\\
    BGRL & \underline{0.3422} & \underline{0.5387} & \underline{0.6597} & \underline{0.8284} & \underline{0.4329} & 0.4821 & 0.7065 & 0.8246 & 0.9243 & 0.6110\\
    GraphMAE & 0.3348 & 0.5298 & 0.6467 & 0.8112 & 0.4147 & 0.4611 & 0.6983 & 0.8167 & 0.9184 & 0.6001 \\
    GraphMAE2 & 0.3287 & 0.5211 & 0.6421 & 0.8019 & 0.4020 & \underline{0.4899} & \textbf{0.7256} & \textbf{0.8389} & \underline{0.9297} & \underline{0.6292} \\
    \midrule
    No-pretrain$_{\small T\hspace{-0.4mm}r\hspace{-0.4mm}a\hspace{-0.4mm}n\hspace{-0.4mm}s\hspace{-0.4mm}f\hspace{-0.4mm}o\hspace{-0.4mm}r\hspace{-0.4mm}m\hspace{-0.4mm}e\hspace{-0.4mm}r}$ & 0.3198 & 0.5101 & 0.6242 & 0.7931 & 0.4001 & 0.4668 & 0.6948 & 0.8119 & 0.9201 & 0.6012 \\
    \model & \textbf{0.3448} & \textbf{0.5429} & \textbf{0.6621} & \textbf{0.8301} & \textbf{0.4420} & \textbf{0.4923} & \underline{0.7189} & \underline{0.8326} & \textbf{0.9319} & \textbf{0.6301}\\
    Improv. vs Best Baseline (\%) & 0.76\% & 0.78\% & 0.36\% & 0.20\% & 2.10\% & 0.49\% &  -0.92\% & -0.75\% & 0.24\% & 0.14\% \\
    \bottomrule[1.1pt]
    \end{tabular}
    }
    \label{tab:flp}
    \vspace{-5mm}
\end{table*}

\vpara{Dataset.} 
We clarify some fundamental concepts in the online gaming business. Tencent's online games and social media platforms are closely interconnected. 
Players can log into various games using their social platform accounts, such as WeChat, and can add new game friends within the games. 
For our pre-training dataset, we select game players on the WeChat platform as nodes and the union of each player's friend relationships in a few games as edges. 
For privacy reasons, we extract a relatively large subgraph from the raw data. 
All subsequently mentioned datasets are constructed in this manner.
We construct a 146-dimensional numerical feature for each node. 
These features include information about the account's login history, activity duration, and in-game item purchase records over a certain period.
Detailed information is shown in Table~\ref{tab:pdata}.

\vpara{Baselines.}
\label{sec:base}
We compare our \model framework with state-of-the-art graph pre-training and self-supervised learning algorithms, DGI~\cite{velivckovic2018deep}, GRACE~\cite{zhu2020deep}, BGRL~\cite{thakoor2021bootstrapped}, GraphMAE~\cite{hou2022graphmae} and GraphMAE2~\cite{hou2023graphmae2}. 
Other methods are not included in the comparison because they are not scalable to large graphs, e.g., GMAE~\cite{zhang2022graph}.
Given the enormous scale of the pre-training graph, sampling algorithms like GraphSAINT~\cite{zeng2019graphsaint} or Neighbor Sampling~\cite{hamilton2017inductive} require more than 72 hours for pre-training.
We adopt ClusterGCN~\cite{chiang2019cluster} as the subgraph sampling method in our baselines due to its superior training efficiency. 
For all baseline methods, we employ GAT~\cite{velivckovic2017graph} as the backbone.

\subsection{Minor Detection Task}
\vpara{Task description.} 
In accordance with relevant national regulations, Tencent's online games have restrictions on the playing time of minors. The objective of the minors detection task is to identify users in a specific online game who appear to be minors.
In this paper, we model it as a binary classification task per node on the graph. For each node to be predicted, we classify it as either an adult or a minor.

\vpara{Setup.}
We built our datasets from the accounts of Game A and B, as illustrated in Table~\ref{tab:pdata}. 
For both datasets, the training, validation, and test sets consist of 100,000, 100,000, and 800,000 nodes, respectively. 

We select all the self-supervised pre-trained models mentioned earlier as baseline methods. 
In alignment with pre-training, we employ ClusterGCN as the subgraph sampling algorithm to scale baseline methods to large graphs.
Following most graph self-supervised learning algorithms~\cite{hou2023graphmae2}, we evaluate our method under two settings: \textit{linear probing} and \textit{fine-tuning}. 
For the \textit{linear probing} setting, 
we freeze the pre-trained encoder and infer node representations. Then, using the node representations as input, we train a linear classifier to obtain the predictions. 
The purpose of the \textit{linear probing} setting is to evaluate the pre-trained model's ability to generate good node representations.
For the \textit{fine-tuning} setting, we attach a linear classifier to the last layer of the pre-trained encoder and then conduct supervised end-to-end fine-tuning. 
The purpose of the \textit{fine-tuning} setting is to evaluate the pre-trained model's ability to transfer its learned knowledge to unseen graphs and downstream tasks in low-label scenarios.
For both settings, we run the experiments 10 times with random seeds.

\vpara{Results.}
Table~\ref{tab:lnc} shows the linear probing results. We observe the following:
(1) \model significantly outperforms all graph self-supervised learning baselines, demonstrating its ability to generate more discriminative representations on unseen graphs.
(2) Some pre-trained models exhibit \textit{negative transfer}, performing worse than the non-pretrained GAT, especially on Game B. This suggests that knowledge learned during pre-training may not effectively transfer to unseen graphs.
(3) Comparing backbones, our graph transformer with PPR sampling significantly outperforms GAT with ClusterGCN sampling. This is likely because ClusterGCN loses over 70-90\% of edges when partitioning large graphs, causing a severe loss of structural information.

Table~\ref{tab:fnc} reports the fine-tuning results. In low-label scenarios, all pre-trained models show improvements compared to the non-pretrained model, demonstrating that pre-training provides better weight initialization for resource-scarce conditions. However, with sufficient labels, some pre-training methods fail to bring consistent improvements, possibly due to overfitting to noise in the training data, which reduces generalization. Across most experimental conditions, \model consistently outperforms all baselines, demonstrating its adaptability to the widely used pretrain-finetune paradigm.

\begin{table}[t]
\vspace{-2mm}
\centering
\caption{\label{tab:apt} Ablation studies of \model pre-training objectives. \textmd{We report ROC-AUC for Game A and B, MRR for Game C and D.}}
   \renewcommand\tabcolsep{3.4pt}
\scalebox{0.8}{
\begin{tabular}{l|cc|cc}
\toprule[1.1pt]
Methods &  Game A  & Game B & Game C  & Game D\\
\midrule 
\model & \textbf{0.7087} & \textbf{0.8121} & \textbf{0.4034} & \textbf{0.6273} \\
w/o feature recon. & 0.6694 & 0.7432 & 0.3525 & 0.5329 \\
w/o local graph recon. & 0.7001 & 0.8056 & 0.3876 & 0.6087\\
\bottomrule[1.1pt]
\end{tabular}
}
\vspace{-4mm}
\end{table}

\begin{table}[t]
\centering
\caption{\label{tab:aif} Ablation studies of key designs in inference. \textmd{We report ROC-AUC for Game A and B, MRR for Game C and D.}}
   \renewcommand\tabcolsep{2pt}
\scalebox{0.8}{
\begin{tabular}{l|cc|cc}
\toprule[1.1pt]
Methods &  Game A  & Game B & Game C  & Game D\\
\midrule 
\model & 0.7087 & \textbf{0.8121} & \textbf{0.4034} & 0.6273 \\
w/ full neighbor sampling & \textbf{0.7111} & 0.8120 & 0.4012 & \textbf{0.6288} \\
w/o proposed feature aug. & 0.6907 & 0.8065 & 0.3921 & 0.6201 \\
\bottomrule[1.1pt]
\end{tabular}
}
\vspace{-4mm}
\end{table}

\subsection{Friend Recall Task}

\vpara{Task description.} 
The friend recall task is designed to encourage active players to invite their friends who have churned, thereby rewarding both parties with certain incentives.
An active player in the game will see a list of their recently inactive friends. They can choose to click and invite them to play together in exchange for specified rewards. When the invitation is sent, the invited player's social media account will receive a message. If the invited player chooses to accept the invitation, they will also receive corresponding rewards.
Essentially, the Friend Recall Task is about ranking churned players among their friends to maximize the probability of successful recall. With the friendships between players, we can model it as a link prediction task on graphs.

\vpara{Setup.}
We build our datasets using friend recall records from Game C and Game D across a span of 5 consecutive months.
The graph for each month can be regarded as a snapshot of friend recall records for a particular game.
If an active player X invites a churned player friend Y, and player Y accepts the invitation, then the edge between them is labeled as a positive edge. 
Conversely, if the churned player Y does not accept the invitation, or if the active player X did not even invite the churned player friend Y, then the edge between them is labeled as negative. 
We train on the first three snapshots, validate on the fourth, and test on the final snapshot.
Table~\ref{tab:fdata} provides a summary of the datasets. The number of edges and the number of nodes are cumulative across the five snapshots.

We select all the self-supervised pre-trained models mentioned earlier as baseline methods. 
For the \textit{linear probing} setting, we still adopt ClusterGCN~\cite{chiang2019cluster} as the sampling algorithm to scale baseline methods to large graphs in inference.
For the ease of mini-batch training on node pairs in the link prediction task, we choose Neighbor Sampling~\cite{hamilton2017inductive} as the sampling algorithm to scale baseline methods to large graphs under \textit{fine-tuning} setting.
We also compared some methods that do not utilize graph structures, such as XGBoost~\cite{chen2016xgboost}, MLP, and Bilinear.
Similar to the node classification task, we conduct experiments under both the \textit{linear probing} and \textit{fine-tuning} settings for self-supervised methods. 
Following the majority of link prediction methods, we randomly sample node pairs within a batch as negative samples to train this classifier.
We use the commonly used ranking metrics Hits@k and Mean Reciprocal Rank (MRR). We run the experiments 10 times with random seeds.

\vpara{Results.}
Table~\ref{tab:llp} shows the results of our method and baseline methods under \textit{linear probing} setting. 
Firstly, on both datasets, \model achieves better results than all self-supervised learning baselines. 
This further underscores that our designed pre-training strategy can bring benefits to various downstream tasks at both the node and edge levels.
Secondly, it can be observed that our method, along with most baseline methods, brings performance improvements over randomly initialized backbone networks through pre-training.

Table~\ref{tab:flp} reports the results under \textit{fine-tuning} setting. 
We can observe that our proposed \model framework consistently outperforms the baseline methods or achieves comparable results on both datasets. However, such improvements are not as significant as under \textit{linear probing} setting. One possible reason is that we adopt Neighbor Sampling to support pairwise mini-batch training for GAT-based algorithms instead of ClusterGCN.
However, the limitation of Neighbor Sampling is that it leads to training inefficiencies when extended to web-scale graphs~\cite{zeng2019graphsaint,chiang2019cluster}. 
These results partially support our argument about the importance of designing consistent and efficient sampling algorithms for both the training and inference phases on web-scale graphs in industrial scenarios.

\begin{figure}[t]
\vspace{-5mm}
    \centering
    \includegraphics[width=1\linewidth]{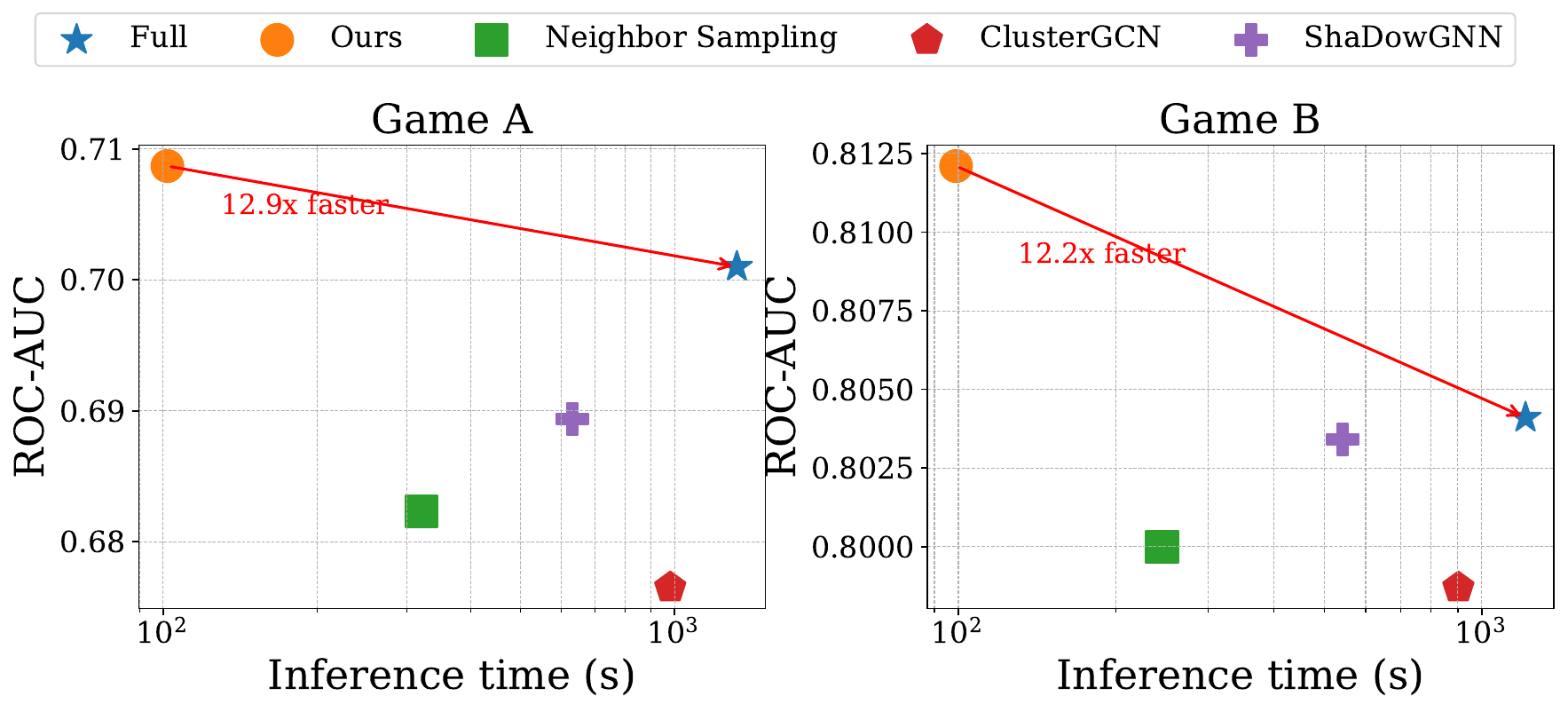}
    \caption{Efficiency analysis of different sampling strategies in inference. \textmd{``Ours" refers to our graph transformer that uses PPR sampling, while the remaining four refer to GAT using conventional sampling strategies.}}
    \label{fig:efficiency}
    \vspace{-3.7mm}
\end{figure}

\begin{table}[t]
\centering
\caption{Comparison of different feature augmentation strategies. We report node classification accuracy (\%).}
\label{tab:feature_aug_ablation}
\scalebox{0.8}{
\begin{tabular}{lccc}
\toprule[1.1pt]
Method & Cora & PubMed & ogbn-arxiv \\
\midrule
PGT w/ proposed feature aug. & \textbf{70.11} & \textbf{69.24} & \textbf{70.22} \\
PGT w/o augmentation & 68.94 & 68.45 & 69.85 \\
PGT w/ feature dropout (10\%) & 68.35 & 67.98 & 69.12 \\
PGT w/ Gaussian noise & 67.92 & 67.53 & 68.77 \\
\bottomrule[1.1pt]
\end{tabular}
}
\vspace{-4mm}
\end{table}

\begin{table}[t]
\centering
\caption{Comparison of different feature augmentation strategies on public datasets. We report node classification accuracy (\%).}
\label{tab:feature_aug}
\scalebox{0.8}{
\begin{tabular}{lccc}
\toprule[1.1pt]
\textbf{Feature Strategy} & \textbf{Cora} & \textbf{PubMed} & \textbf{ogbn-arxiv} \\
\midrule
Original only & 68.94$\pm$0.38 & 68.45$\pm$0.87 & 69.85$\pm$0.37 \\
Reconstructed only & 66.84$\pm$0.46 & 65.28$\pm$0.94 & 68.75$\pm$0.42 \\
Average pooling & 70.11$\pm$0.44 & \textbf{69.24$\pm$1.21} & \textbf{70.22$\pm$0.42} \\
Sum pooling & \textbf{70.14$\pm$0.51} & 69.18$\pm$0.89 & 70.09$\pm$0.38 \\
\bottomrule[1.1pt]
\end{tabular}
}
\vspace{-4mm}
\end{table}

\begin{table*}[!t]
\vspace{-7mm}
\centering
\caption{PPR computation time for different datasets.}
\label{tab:ppr_time}
\scalebox{0.8}{
\begin{tabular}{lrrr}
\toprule[1.1pt]
Dataset & Number of Seed Nodes & PPR Computation Time & Infrastructure \\
\midrule
Pre-training Graph & 54M (10\% of all nodes) & $\sim$8 hours & Spark+Angel (200 workers) \\
Game A & 1M (labeled nodes) & $\sim$30 minutes & Spark+Angel (200 workers) \\
Game B & 1M (labeled nodes) & $\sim$25 minutes & Spark+Angel (200 workers) \\
Game C & 0.7M (labeled nodes) & $\sim$20 minutes & Spark+Angel (200 workers) \\
Game D & 0.5M (labeled nodes) & $\sim$15 minutes & Spark+Angel (200 workers) \\
ogbn-papers100M & 11M (10\% of all nodes) & $\sim$2 hours & Single machine (128 cores) \\
ogbn-arxiv & 0.17M (all nodes) & $\sim$10 minutes & Single machine (128 cores) \\
\bottomrule[1.1pt]
\end{tabular}
}
\vspace{-5mm}
\end{table*}

\begin{table}[!t]
\caption{Sensitivity analysis of mixing coefficient $\lambda$.}
\label{tab:lambda_sensitivity}
\centering
\scalebox{0.8}{
\begin{tabular}{c|ccc}
\toprule[1.1pt]
$\lambda$ Value & Cora & PubMed & ogbn-arxiv \\
\midrule
0.01 & $69.78 \pm 0.52$ & $68.92 \pm 1.32$ & $70.01 \pm 0.57$ \\
0.05 & \textbf{70.14 $\pm$ 0.48} & $69.09 \pm 1.27$ & $69.87 \pm 0.45$ \\
0.1 & $70.11 \pm 0.44$ & \textbf{69.24 $\pm$ 1.21} & \textbf{70.22 $\pm$ 0.42} \\
0.2 & $70.06 \pm 0.51$ & $69.17 \pm 1.25$ & $70.13 \pm 0.49$ \\
0.5 & $70.10 \pm 0.57$ & $68.89 \pm 1.31$ & $69.95 \pm 0.54$ \\
\bottomrule[1.1pt]
\end{tabular}
}
\vspace{-4mm}
\end{table}

\begin{table}[t]
\caption{Sensitivity analysis of mask rate.}
\label{tab:mask_rate_sensitivity}
\centering
\scalebox{0.8}{
\begin{tabular}{cccc}
\toprule[1.1pt]
Mask Rate & Cora & PubMed & ogbn-arxiv \\
\midrule
0.65 & 69.32 $\pm$ 0.63 & 68.57 $\pm$ 1.33 & 69.42 $\pm$ 0.58 \\
0.75 & 69.78 $\pm$ 0.51 & 68.89 $\pm$ 1.29 & 69.85 $\pm$ 0.49 \\
0.85 & \textbf{70.11 $\pm$ 0.44} & \textbf{69.24 $\pm$ 1.21} & \textbf{70.22 $\pm$ 0.42}\\
0.90 & 69.87 $\pm$ 0.49 & 69.01 $\pm$ 1.27 & 69.96 $\pm$ 0.47 \\
0.95 & 69.15 $\pm$ 0.61 & 68.43 $\pm$ 1.36 & 69.27 $\pm$ 0.56 \\
\bottomrule[1.1pt]
\end{tabular}
}
\vspace{-5mm}
\end{table}

\begin{table}[t]
\caption{Sensitivity analysis of PPR topk.}
\label{tab:ppr_topk_sensitivity}
\centering
\scalebox{0.8}{
\begin{tabular}{cccc}
\toprule[1.1pt]
PPR topk & Cora & PubMed & ogbn-arxiv \\
\midrule
64 & 69.54 $\pm$ 0.53 & 68.76 $\pm$ 1.31 & 69.67 $\pm$ 0.55 \\
96 & 69.89 $\pm$ 0.48 & 69.05 $\pm$ 1.25 & 69.97 $\pm$ 0.46 \\
128 & \textbf{70.11 $\pm$ 0.44} & \textbf{69.24 $\pm$ 1.21} & \textbf{70.22 $\pm$ 0.42}\\
160 & 70.08 $\pm$ 0.47 & 69.17 $\pm$ 1.24 & 70.15 $\pm$ 0.45 \\
192 & 69.92 $\pm$ 0.51 & 68.98 $\pm$ 1.28 & 70.01 $\pm$ 0.48 \\
\bottomrule[1.1pt]
\end{tabular}
}
\vspace{-4mm}
\end{table}

\subsection{Ablation Studies}
We conduct a series of ablation studies to verify the contributions of the designs in \model under the \textit{linear probing} setting.

\vpara{Effect of pre-training objectives.}
Table~\ref{tab:apt} shows the results, where ``w/o feature recon." represents that we only employ the local graph structure reconstruction objective in pre-training. Conversely, ``w/o local graph recon." signifies that only the feature reconstruction objective is used. 
It can be observed that for two downstream tasks, both proposed pre-training objectives contribute to performance. Comparatively, the removal of the feature reconstruction objective results in a larger performance degradation, implying the significance of node features for graph learning tasks. 

\vpara{Effect of key designs in inference.}
Table~\ref{tab:aif} shows the influence of key proposed designs in inference. 
Our proposed method of employing PPR sampling for inference attains performance comparable to that achieved using all neighbor nodes. This substantiates the efficacy of PPR in selecting informative nodes without the necessity for a large receptive field.
The notation ``w/o proposed feature aug." represents the absence of our proposed strategy of reusing the decoder for feature augmentation. 
Functioning as a plug-and-play component, it effectively improves performance across all datasets. 

\vpara{Efficiency analysis.}
Fig.~\ref{fig:efficiency} compares inference performance and time of different sampling methods, using the same number of nodes to be predicted (test set on each dataset) and the same hardware conditions. 
It can be observed that our method delivers the most favorable outcomes in both performance and speed. 
It is noteworthy that, in contrast to the efficiency of ClusterGCN sampling during the training phase, it does not exhibit advantages in inference. 
A possible explanation could be that the nodes requiring predictions constitute only a small fraction of all nodes. 
The GNN still needs to perform inference across all sub-graphs to obtain all prediction results. 
This further substantiates the significance of decoupling the sampling of labelled and auxiliary nodes.

\subsection{Ablation Study on Feature Augmentation Strategies}
To validate our proposed feature augmentation strategy, we compare it against several alternatives: no augmentation (original features only), feature dropout (10\%), and adding Gaussian noise ($\mathcal{N}(0, 0.1)$). As shown in Table~\ref{tab:feature_aug_ablation}, our decoder-based augmentation consistently outperforms all other strategies. While common techniques like feature dropout and adding noise provide less benefit and can harm performance, our method yields consistent and meaningful improvements.

The effectiveness of our approach stems from its ability to leverage patterns learned during pre-training. Unlike random perturbations, the decoder generates features that are semantically consistent with the originals, acting as a form of "learned feature denoising." This process enhances feature representations by incorporating information from the graph's global structure, confirming that the decoder captures meaningful patterns that benefit downstream tasks.

We further analyze how to best combine the original features ($X$) with the reconstructed features ($X'$). We evaluate four strategies:
\begin{itemize}
    \item \textbf{Original features only:} Using only the original node features $X$.
    \item \textbf{Reconstructed features only:} Using only the decoder-reconstructed features $X'$.
    \item \textbf{Average pooling:} Using $\frac{1}{2}(X + X')$ (our proposed method).
    \item \textbf{Sum pooling:} Using $(X + X')$ without normalization.
\end{itemize}

Table~\ref{tab:feature_aug} shows that combining features via average or sum pooling significantly outperforms using either original or reconstructed features alone. This confirms that the two feature sets provide complementary information, leading to more robust representations.

The performance of average and sum pooling is statistically similar. Theoretically, average pooling maintains the original feature scale, which can stabilize training. Sum pooling doubles the feature magnitude, which could require re-tuning other parameters. We adopt average pooling as our default strategy because it preserves the feature scale and performed slightly better in our experiments, though both are effective combination methods.

\subsection{Validating The Choice of Transformers as The Backbone Architecture}
To validate our choice of a transformer backbone, we compare it against a GNN-based approach across all experiments. As shown in Tables~\ref{tab:pnc}, \ref{tab:diverse_domains}, \ref{tab:lnc}, \ref{tab:fnc}, and \ref{tab:flp}, the non-pretrained transformer consistently outperforms its GNN counterpart (No-pretrainGAT), even without pre-training. For instance, on the minors detection task (Table~\ref{tab:lnc}), the transformer baseline achieves ROC-AUC scores of 0.6771 and 0.8021 on Games A and B respectively, compared to 0.6621 and 0.7829 for the GAT baseline. This performance gap widens further after applying our pre-training framework. 

We attribute this superiority to the transformer's ability to: (1) handle heterophilic patterns common in industrial data; (2) remain robust to structural noise via self-attention; and (3) capture long-range dependencies without the over-smoothing issues that plague deep GNNs. Furthermore, as illustrated in Fig.~\ref{fig:efficiency}, our transformer-based approach with PPR sampling is up to 12.9x faster in inference than full neighborhood aggregation methods while maintaining superior performance, addressing a critical requirement for industrial deployment.

\subsection{Time Complexity Analysis}
The complexity of our method comprises three main components: PPR computation, model training, and inference. For PPR computation, we use the efficient Forward Push algorithm with a complexity of approximately $O(k/\varepsilon)$ per seed node, which is far more scalable than the $O(|V|^3)$ complexity of exact PPR. During training and inference, our model operates on fixed-length sequences. The training complexity for a batch of $b$ nodes is $O(b \times s^2 \times d)$, where $s$ is the sequence length, and inference is $O(s^2 \times d)$ per node. This is significantly more efficient than full-graph attention ($O(|V|^2 \times d)$) or recursive neighborhood expansion.

Table~\ref{tab:ppr_time} details the practical computation times. For our 540M-node pre-training graph, computing PPR sequences for 54M seed nodes took approximately 8 hours on a 200-worker distributed cluster. On downstream tasks, this process took only 15-30 minutes. These measurements confirm the practicality of our approach for industrial applications requiring both training efficiency and low-latency inference.

\subsection{Hyperparameter Sensitivity Analysis}
To provide a more comprehensive understanding of our framework's behavior, we conducted sensitivity analyses on three key hyperparameters: the mixing coefficient $\lambda$, mask rate, and PPR topk. We performed these experiments on three public datasets: Cora, PubMed, and ogbn-arxiv.

\vpara{Sensitivity Analysis of $\lambda$.}
We varied $\lambda$ across the values ${0.01, 0.05, 0.1, 0.2, 0.5}$ to understand its impact on model performance.
The results are presented in Table~\ref{tab:lambda_sensitivity}.
These results demonstrate that while $\lambda = 0.1$ yields optimal performance across all datasets, the model's performance does not vary significantly across different values of $\lambda$. The maximum performance variation is approximately 0.4 percentage points across the tested range, indicating that our framework is not highly sensitive to this hyperparameter.
This robustness to the $\lambda$ setting is advantageous from a practical perspective, as it suggests that precise tuning of this hyperparameter is not critical for achieving good performance with our framework across different datasets.

\vpara{Sensitivity Analysis of Mask Rate.}
We varied the mask rate across the values {0.65, 0.75, 0.85, 0.90, 0.95} to evaluate its effect on performance. Table~\ref{tab:mask_rate_sensitivity} presents the results.
The results indicate that a mask rate of 0.85 yields the best performance across all datasets. This is consistent with findings in other masked modeling works that high mask rates (but not too extreme) often lead to optimal performance by forcing the model to learn more robust representations.

\vpara{Sensitivity Analysis of PPR topk.}
We varied the PPR topk parameter across the values {64, 96, 128, 160, 192} to understand its impact on the model's performance. Table~\ref{tab:ppr_topk_sensitivity} shows the results.
The results show that a PPR topk value of 128 consistently provides the best performance across all datasets. The model's performance is relatively stable across different topk values, with only minor fluctuations.

%% file: 5.conclusion.tex
\section{conclusion}

In this work, we explore the application of graph pre-training in the online gaming industry. 
We propose a novel, scalable Pre-trained Graph Transformer (PGT) tailored for web-scale graphs, specifically in the online gaming industry. 
Empirical results on both real industrial data at Tencent and public benchmarks substantiate the model’s state-of-the-art performance and exceptional generalizability across different graphs and tasks. Our experiments on dynamic link prediction further underscore this versatility, showing that a pre-trained static encoder can be a superior foundation for temporal tasks compared to training specialized dynamic models from scratch.
In future work, we plan to further broaden the scope, efficiency, and interpretability of graph pre-trained models, with the ultimate aim of contributing to the establishment of a graph foundation model. 
The success of PGT on both static and dynamic tasks represents a significant step in this direction, suggesting a promising path towards unified models for diverse graph learning problems.

%% file: reference.bib
@article{hamilton2017inductive,
  title={Inductive representation learning on large graphs},
  author={Hamilton, Will and Ying, Zhitao and Leskovec, Jure},
  journal={NeurIPS},
  volume={30},
  year={2017}
}

@article{kipf2016semi,
  title={Semi-supervised classification with graph convolutional networks},
  author={Kipf, Thomas N and Welling, Max},
  journal={ICLR},
  year={2016}
}

@article{velivckovic2017graph,
  title={Graph attention networks},
  author={Veli{\v{c}}kovi{\'c}, Petar and Cucurull, Guillem and Casanova, Arantxa and Romero, Adriana and Lio, Pietro and Bengio, Yoshua},
  journal={ICLR},
  year={2017}
}

@article{xu2018powerful,
  title={How powerful are graph neural networks?},
  author={Xu, Keyulu and Hu, Weihua and Leskovec, Jure and Jegelka, Stefanie},
  journal={ICLR},
  year={2018}
}

@inproceedings{chiang2019cluster,
  title={Cluster-gcn: An efficient algorithm for training deep and large graph convolutional networks},
  author={Chiang, Wei-Lin and Liu, Xuanqing and Si, Si and Li, Yang and Bengio, Samy and Hsieh, Cho-Jui},
  booktitle={KDD},
  pages={257--266},
  year={2019}
}

@article{zeng2019graphsaint,
  title={Graphsaint: Graph sampling based inductive learning method},
  author={Zeng, Hanqing and Zhou, Hongkuan and Srivastava, Ajitesh and Kannan, Rajgopal and Prasanna, Viktor},
  journal={ICLR},
  year={2019}
}

@article{vaswani2017attention,
  title={Attention is all you need},
  author={Vaswani, Ashish and Shazeer, Noam and Parmar, Niki and Uszkoreit, Jakob and Jones, Llion and Gomez, Aidan N and Kaiser, {\L}ukasz and Polosukhin, Illia},
  journal={NeurIPS},
  volume={30},
  year={2017}
}

@article{dosovitskiy2020image,
  title={An image is worth 16x16 words: Transformers for image recognition at scale},
  author={Dosovitskiy, Alexey and Beyer, Lucas and Kolesnikov, Alexander and Weissenborn, Dirk and Zhai, Xiaohua and Unterthiner, Thomas and Dehghani, Mostafa and Minderer, Matthias and Heigold, Georg and Gelly, Sylvain and others},
  journal={ICLR},
  year={2021}
}

@article{dwivedi2020generalization,
  title={A generalization of transformer networks to graphs},
  author={Dwivedi, Vijay Prakash and Bresson, Xavier},
  journal={arXiv preprint arXiv:2012.09699},
  year={2020}
}

@article{zeng2021decoupling,
  title={Decoupling the depth and scope of graph neural networks},
  author={Zeng, Hanqing and Zhang, Muhan and Xia, Yinglong and Srivastava, Ajitesh and Malevich, Andrey and Kannan, Rajgopal and Prasanna, Viktor and Jin, Long and Chen, Ren},
  journal={NeurIPS},
  volume={34},
  pages={19665--19679},
  year={2021}
}

@inproceedings{chen2021litegt,
  title={Litegt: Efficient and lightweight graph transformers},
  author={Chen, Cong and Tao, Chaofan and Wong, Ngai},
  booktitle={CIKM},
  pages={161--170},
  year={2021}
}

@inproceedings{chen2022structure,
  title={Structure-aware transformer for graph representation learning},
  author={Chen, Dexiong and O’Bray, Leslie and Borgwardt, Karsten},
  booktitle={ICML},
  pages={3469--3489},
  year={2022},
  organization={PMLR}
}

@article{ying2021transformers,
  title={Do transformers really perform badly for graph representation?},
  author={Ying, Chengxuan and Cai, Tianle and Luo, Shengjie and Zheng, Shuxin and Ke, Guolin and He, Di and Shen, Yanming and Liu, Tie-Yan},
  journal={NeurIPS},
  volume={34},
  pages={28877--28888},
  year={2021}
}

@article{masters2022gps++,
  title={Gps++: An optimised hybrid mpnn/transformer for molecular property prediction},
  author={Masters, Dominic and Dean, Josef and Klaser, Kerstin and Li, Zhiyi and Maddrell-Mander, Sam and Sanders, Adam and Helal, Hatem and Beker, Deniz and Ramp{\'a}{\v{s}}ek, Ladislav and Beaini, Dominique},
  journal={ICLR},
  year={2022}
}

@article{hu2021ogb,
  title={Ogb-lsc: A large-scale challenge for machine learning on graphs},
  author={Hu, Weihua and Fey, Matthias and Ren, Hongyu and Nakata, Maho and Dong, Yuxiao and Leskovec, Jure},
  journal={arXiv preprint arXiv:2103.09430},
  year={2021}
}

@article{zhao2021gophormer,
  title={Gophormer: Ego-graph transformer for node classification},
  author={Zhao, Jianan and Li, Chaozhuo and Wen, Qianlong and Wang, Yiqi and Liu, Yuming and Sun, Hao and Xie, Xing and Ye, Yanfang},
  journal={arXiv preprint arXiv:2110.13094},
  year={2021}
}

@article{wu2022nodeformer,
  title={Nodeformer: A scalable graph structure learning transformer for node classification},
  author={Wu, Qitian and Zhao, Wentao and Li, Zenan and Wipf, David P and Yan, Junchi},
  journal={NeurIPS},
  volume={35},
  pages={27387--27401},
  year={2022}
}

@article{zhang2022hierarchical,
  title={Hierarchical graph transformer with adaptive node sampling},
  author={Zhang, Zaixi and Liu, Qi and Hu, Qingyong and Lee, Chee-Kong},
  journal={NeurIPS},
  volume={35},
  pages={21171--21183},
  year={2022}
}

@article{devlin2018bert,
  title={Bert: Pre-training of deep bidirectional transformers for language understanding},
  author={Devlin, Jacob and Chang, Ming-Wei and Lee, Kenton and Toutanova, Kristina},
  journal={NAACL},
  year={2018}
}

@article{hu2019strategies,
  title={Strategies for pre-training graph neural networks},
  author={Hu, Weihua and Liu, Bowen and Gomes, Joseph and Zitnik, Marinka and Liang, Percy and Pande, Vijay and Leskovec, Jure},
  journal={ICLR},
  year={2019}
}

@article{velivckovic2018deep,
  title={Deep graph infomax},
  author={Veli{\v{c}}kovi{\'c}, Petar and Fedus, William and Hamilton, William L and Li{\`o}, Pietro and Bengio, Yoshua and Hjelm, R Devon},
  journal={ICLR},
  year={2018}
}

@inproceedings{hassani2020contrastive,
  title={Contrastive multi-view representation learning on graphs},
  author={Hassani, Kaveh and Khasahmadi, Amir Hosein},
  booktitle={ICML},
  pages={4116--4126},
  year={2020},
  organization={PMLR}
}

@article{zhu2020deep,
  title={Deep graph contrastive representation learning},
  author={Zhu, Yanqiao and Xu, Yichen and Yu, Feng and Liu, Qiang and Wu, Shu and Wang, Liang},
  journal={arXiv preprint arXiv:2006.04131},
  year={2020}
}

@article{grill2020bootstrap,
  title={Bootstrap your own latent-a new approach to self-supervised learning},
  author={Grill, Jean-Bastien and Strub, Florian and Altch{\'e}, Florent and Tallec, Corentin and Richemond, Pierre and Buchatskaya, Elena and Doersch, Carl and Avila Pires, Bernardo and Guo, Zhaohan and Gheshlaghi Azar, Mohammad and others},
  journal={NeurIPS},
  volume={33},
  pages={21271--21284},
  year={2020}
}

@inproceedings{thakoor2021bootstrapped,
  title={Bootstrapped representation learning on graphs},
  author={Thakoor, Shantanu and Tallec, Corentin and Azar, Mohammad Gheshlaghi and Munos, R{\'e}mi and Veli{\v{c}}kovi{\'c}, Petar and Valko, Michal},
  booktitle={ICLR 2021 Workshop on Geometrical and Topological Representation Learning},
  year={2021}
}

@article{kipf2016variational,
  title={Variational graph auto-encoders},
  author={Kipf, Thomas N and Welling, Max},
  journal={arXiv preprint arXiv:1611.07308},
  year={2016}
}

@inproceedings{wang2017mgae,
  title={Mgae: Marginalized graph autoencoder for graph clustering},
  author={Wang, Chun and Pan, Shirui and Long, Guodong and Zhu, Xingquan and Jiang, Jing},
  booktitle={CIKM},
  pages={889--898},
  year={2017}
}

@inproceedings{hou2022graphmae,
  title={Graphmae: Self-supervised masked graph autoencoders},
  author={Hou, Zhenyu and Liu, Xiao and Cen, Yukuo and Dong, Yuxiao and Yang, Hongxia and Wang, Chunjie and Tang, Jie},
  booktitle={KDD},
  pages={594--604},
  year={2022}
}

@inproceedings{hou2023graphmae2,
  title={GraphMAE2: A Decoding-Enhanced Masked Self-Supervised Graph Learner},
  author={Hou, Zhenyu and He, Yufei and Cen, Yukuo and Liu, Xiao and Dong, Yuxiao and Kharlamov, Evgeny and Tang, Jie},
  booktitle={WWW},
  pages={737--746},
  year={2023}
}

@article{li2022maskgae,
  title={Maskgae: Masked graph modeling meets graph autoencoders},
  author={Li, Jintang and Wu, Ruofan and Sun, Wangbin and Chen, Liang and Tian, Sheng and Zhu, Liang and Meng, Changhua and Zheng, Zibin and Wang, Weiqiang},
  journal={KDD},
  year={2023}
}

@inproceedings{qiu2020gcc,
  title={Gcc: Graph contrastive coding for graph neural network pre-training},
  author={Qiu, Jiezhong and Chen, Qibin and Dong, Yuxiao and Zhang, Jing and Yang, Hongxia and Ding, Ming and Wang, Kuansan and Tang, Jie},
  booktitle={KDD},
  pages={1150--1160},
  year={2020}
}

@article{cao2023pre,
  title={When to Pre-Train Graph Neural Networks? An Answer from Data Generation Perspective!},
  author={Cao, Yuxuan and Xu, Jiarong and Yang, Carl and Wang, Jiaan and Zhang, Yunchao and Wang, Chunping and Chen, Lei and Yang, Yang},
  journal={KDD},
  year={2023}
}

@inproceedings{chen2016xgboost,
  title={Xgboost: A scalable tree boosting system},
  author={Chen, Tianqi and Guestrin, Carlos},
  booktitle={KDD},
  pages={785--794},
  year={2016}
}

@inproceedings{he2022masked,
  title={Masked autoencoders are scalable vision learners},
  author={He, Kaiming and Chen, Xinlei and Xie, Saining and Li, Yanghao and Doll{\'a}r, Piotr and Girshick, Ross},
  booktitle={CVPR},
  pages={16000--16009},
  year={2022}
}

@inproceedings{vincent2008extracting,
  title={Extracting and composing robust features with denoising autoencoders},
  author={Vincent, Pascal and Larochelle, Hugo and Bengio, Yoshua and Manzagol, Pierre-Antoine},
  booktitle={ICML},
  pages={1096--1103},
  year={2008}
}

@inproceedings{guo2019attention,
  title={Attention based spatial-temporal graph convolutional networks for traffic flow forecasting},
  author={Guo, Shengnan and Lin, Youfang and Feng, Ning and Song, Chao and Wan, Huaiyu},
  booktitle={AAAI},
  volume={33},
  number={01},
  pages={922--929},
  year={2019}
}

@article{wang2021review,
  title={A review on graph neural network methods in financial applications},
  author={Wang, Jianian and Zhang, Sheng and Xiao, Yanghua and Song, Rui},
  journal={arXiv preprint arXiv:2111.15367},
  year={2021}
}

@inproceedings{ying2018graph,
  title={Graph convolutional neural networks for web-scale recommender systems},
  author={Ying, Rex and He, Ruining and Chen, Kaifeng and Eksombatchai, Pong and Hamilton, William L and Leskovec, Jure},
  booktitle={KDD},
  pages={974--983},
  year={2018}
}

@article{allamanis2022graph,
  title={Graph neural networks in program analysis},
  author={Allamanis, Miltiadis},
  journal={Graph neural networks: foundations, frontiers, and applications},
  pages={483--497},
  year={2022},
  publisher={Springer}
}

@article{wang2019deep,
  title={Deep graph library: A graph-centric, highly-performant package for graph neural networks},
  author={Wang, Minjie and Zheng, Da and Ye, Zihao and Gan, Quan and Li, Mufei and Song, Xiang and Zhou, Jinjing and Ma, Chao and Yu, Lingfan and Gai, Yu and others},
  journal={arXiv preprint arXiv:1909.01315},
  year={2019}
}

@inproceedings{zhang2023inferturbo,
  title={InferTurbo: A Scalable System for Boosting Full-graph Inference of Graph Neural Network over Huge Graphs},
  author={Zhang, Dalong and Song, Xianzheng and Hu, Zhiyang and Li, Yang and Tao, Miao and Hu, Binbin and Wang, Lin and Zhang, Zhiqiang and Zhou, Jun},
  booktitle={ICDE},
  pages={3235--3247},
  year={2023},
  organization={IEEE}
}

@inproceedings{radford2021learning,
  title={Learning transferable visual models from natural language supervision},
  author={Radford, Alec and Kim, Jong Wook and Hallacy, Chris and Ramesh, Aditya and Goh, Gabriel and Agarwal, Sandhini and Sastry, Girish and Askell, Amanda and Mishkin, Pamela and Clark, Jack and others},
  booktitle={ICML},
  pages={8748--8763},
  year={2021},
  organization={PMLR}
}

@article{brown2020language,
  title={Language models are few-shot learners},
  author={Brown, Tom and Mann, Benjamin and Ryder, Nick and Subbiah, Melanie and Kaplan, Jared D and Dhariwal, Prafulla and Neelakantan, Arvind and Shyam, Pranav and Sastry, Girish and Askell, Amanda and others},
  journal={NeurIPS},
  volume={33},
  pages={1877--1901},
  year={2020}
}

@article{fox2019robust,
  title={How robust are graph neural networks to structural noise?},
  author={Fox, James and Rajamanickam, Sivasankaran},
  journal={arXiv preprint arXiv:1912.10206},
  year={2019}
}

@article{zheng2022graph,
  title={Graph neural networks for graphs with heterophily: A survey},
  author={Zheng, Xin and Liu, Yixin and Pan, Shirui and Zhang, Miao and Jin, Di and Yu, Philip S},
  journal={arXiv preprint arXiv:2202.07082},
  year={2022}
}

@article{mccallum2000automating,
  title={Automating the construction of internet portals with machine learning},
  author={McCallum, Andrew Kachites and Nigam, Kamal and Rennie, Jason and Seymore, Kristie},
  journal={Information Retrieval},
  volume={3},
  pages={127--163},
  year={2000},
  publisher={Springer}
}

@article{sen2008collective,
  title={Collective classification in network data},
  author={Sen, Prithviraj and Namata, Galileo and Bilgic, Mustafa and Getoor, Lise and Galligher, Brian and Eliassi-Rad, Tina},
  journal={AI magazine},
  volume={29},
  number={3},
  pages={93--93},
  year={2008}
}

@article{he2023explanations,
  title={Explanations as Features: LLM-Based Features for Text-Attributed Graphs},
  author={He, Xiaoxin and Bresson, Xavier and Laurent, Thomas and Hooi, Bryan},
  journal={arXiv preprint arXiv:2305.19523},
  year={2023}
}

@inproceedings{yang2022large,
  title={Large-scale personalized video game recommendation via social-aware contextualized graph neural network},
  author={Yang, Liangwei and Liu, Zhiwei and Wang, Yu and Wang, Chen and Fan, Ziwei and Yu, Philip S},
  booktitle={WWW},
  pages={3376--3386},
  year={2022}
}

@article{zhang2022graph,
  title={Graph masked autoencoders with transformers},
  author={Zhang, Sixiao and Chen, Hongxu and Yang, Haoran and Sun, Xiangguo and Yu, Philip S and Xu, Guandong},
  journal={arXiv preprint arXiv:2202.08391},
  year={2022}
}

@article{green2021statistical,
  title={Statistical guarantees for local spectral clustering on random neighborhood graphs},
  author={Green, Alden and Balakrishnan, Sivaraman and Tibshirani, Ryan J},
  journal={JMLR},
  volume={22},
  number={1},
  pages={11184--11254},
  year={2021},
  publisher={JMLRORG}
}

@article{tremblay2020approximating,
  title={Approximating spectral clustering via sampling: a review},
  author={Tremblay, Nicolas and Loukas, Andreas},
  journal={Sampling Techniques for Supervised or Unsupervised Tasks},
  pages={129--183},
  year={2020},
  publisher={Springer}
}

@article{vincent2010stacked,
  title={Stacked denoising autoencoders: Learning useful representations in a deep network with a local denoising criterion.},
  author={Vincent, Pascal and Larochelle, Hugo and Lajoie, Isabelle and Bengio, Yoshua and Manzagol, Pierre-Antoine and Bottou, L{\'e}on},
  journal={JMLR},
  volume={11},
  number={12},
  year={2010}
}

@article{mikolov2013distributed,
  title={Distributed representations of words and phrases and their compositionality},
  author={Mikolov, Tomas and Sutskever, Ilya and Chen, Kai and Corrado, Greg S and Dean, Jeff},
  journal={NeurIPS},
  volume={26},
  year={2013}
}

@article{liao2018attributed,
  title={Attributed social network embedding},
  author={Liao, Lizi and He, Xiangnan and Zhang, Hanwang and Chua, Tat-Seng},
  journal={TKDE},
  volume={30},
  number={12},
  pages={2257--2270},
  year={2018},
  publisher={IEEE}
}

@inproceedings{wu2019simplifying,
  title={Simplifying graph convolutional networks},
  author={Wu, Felix and Souza, Amauri and Zhang, Tianyi and Fifty, Christopher and Yu, Tao and Weinberger, Kilian},
  booktitle={ICML},
  pages={6861--6871},
  year={2019},
  organization={Pmlr}
}

@article{ma2025acceleration,
  title={Acceleration algorithms in gnns: A survey},
  author={Ma, Lu and Sheng, Zeang and Li, Xunkai and Gao, Xinyi and Hao, Zhezheng and Yang, Ling and Nie, Xiaonan and Jiang, Jiawei and Zhang, Wentao and Cui, Bin},
  journal={TKDE},
  year={2025},
  publisher={IEEE}
}

@inproceedings{pareja2020evolvegcn,
  title={Evolvegcn: Evolving graph convolutional networks for dynamic graphs},
  author={Pareja, Aldo and Domeniconi, Giacomo and Chen, Jie and Ma, Tengfei and Suzumura, Toyotaro and Kanezashi, Hiroki and Kaler, Tim and Schardl, Tao B and Leiserson, Charles E},
  booktitle={Proceedings of the AAAI conference on artificial intelligence},
  volume={34},
  number={04},
  pages={5363--5370},
  year={2020}
}

@inproceedings{rossi2020temporal,
  title={Temporal graph networks for deep learning on dynamic graphs},
  author={Rossi, Emanuele and Chamberlain, Ben and Frasca, Fabrizio and Hamilton, William L and Lio, Pietro and Bronstein, Michael M},
  booktitle={ICLR 2020 Workshop on Deep Learning on Graphs},
  year={2020}
}

@article{gao2022novel,
  title={A novel representation learning for dynamic graphs based on graph convolutional networks},
  author={Gao, Chao and Zhu, Junyou and Zhang, Fan and Wang, Zhen and Li, Xuelong},
  journal={IEEE Transactions on Cybernetics},
  volume={53},
  number={6},
  pages={3599--3612},
  year={2022},
  publisher={IEEE}
}

@inproceedings{hajiramezanali2019variational,
  title={Variational graph recurrent neural networks},
  author={Hajiramezanali, Ehsan and Hasanzadeh, Arman and Ghanbari, Morteza and Ghafari, Mahdi and Duffield, Nick and Narayanan, Krishna R},
  booktitle={Advances in Neural Information Processing Systems},
  volume={32},
  year={2019}
}

@article{chung2014empirical,
  title={Empirical evaluation of gated recurrent neural networks on sequence modeling},
  author={Chung, Junyoung and Gulcehre, Caglar and Cho, KyungHyun and Bengio, Yoshua},
  journal={arXiv preprint arXiv:1412.3555},
  year={2014}
}

@inproceedings{DBLP:conf/aaai/0016HHW25,
  author       = {Jun Hu and
                  Bryan Hooi and
                  Bingsheng He and
                  Yinwei Wei},
  editor       = {Toby Walsh and
                  Julie Shah and
                  Zico Kolter},
  title        = {Modality-Independent Graph Neural Networks with Global Transformers
                  for Multimodal Recommendation},
  booktitle    = {AAAI-25, Sponsored by the Association for the Advancement of Artificial
                  Intelligence, February 25 - March 4, 2025, Philadelphia, PA, {USA}},
  pages        = {11790--11798},
  publisher    = {{AAAI} Press},
  year         = {2025},
  url          = {https://doi.org/10.1609/aaai.v39i11.33283},
  doi          = {10.1609/AAAI.V39I11.33283},
  bibsource    = {dblp computer science bibliography, https://dblp.org}
}

@article{DBLP:journals/tkde/HuHQFX24,
  author       = {Jun Hu and
                  Bryan Hooi and
                  Shengsheng Qian and
                  Quan Fang and
                  Changsheng Xu},
  title        = {{MGDCF:} Distance Learning via Markov Graph Diffusion for Neural Collaborative
                  Filtering},
  journal      = {{IEEE} Trans. Knowl. Data Eng.},
  volume       = {36},
  number       = {7},
  pages        = {3281--3296},
  year         = {2024},
  url          = {https://doi.org/10.1109/TKDE.2023.3348537},
  doi          = {10.1109/TKDE.2023.3348537},
  bibsource    = {dblp computer science bibliography, https://dblp.org}
}

@inproceedings{he2025unigraph,
  title={UniGraph: Learning a Unified Cross-Domain Foundation Model for Text-Attributed Graphs},
  author={He, Yufei and Sui, Yuan and He, Xiaoxin and Hooi, Bryan},
  booktitle={Proceedings of the 31st ACM SIGKDD Conference on Knowledge Discovery and Data Mining V. 1},
  pages={448--459},
  year={2025}
}

@inproceedings{he2025unigraph2,
  title={Unigraph2: Learning a unified embedding space to bind multimodal graphs},
  author={He, Yufei and Sui, Yuan and He, Xiaoxin and Liu, Yue and Sun, Yifei and Hooi, Bryan},
  booktitle={Proceedings of the ACM on Web Conference 2025},
  pages={1759--1770},
  year={2025}
}

@article{sui2024fidelis,
  title={Fidelis: Faithful reasoning in large language model for knowledge graph question answering},
  author={Sui, Yuan and He, Yufei and Liu, Nian and He, Xiaoxin and Wang, Kun and Hooi, Bryan},
  journal={ACL},
  year={2025}
}

@article{sui2024can,
  title={Can knowledge graphs make large language models more trustworthy? an empirical study over open-ended question answering},
  author={Sui, Yuan and He, Yufei and Ding, Zifeng and Hooi, Bryan},
  journal={ACL},
  year={2025}
}

@article{liu2025guardreasoner,
  title={Guardreasoner: Towards reasoning-based llm safeguards},
  author={Liu, Yue and Gao, Hongcheng and Zhai, Shengfang and Xia, Jun and Wu, Tianyi and Xue, Zhiwei and Chen, Yulin and Kawaguchi, Kenji and Zhang, Jiaheng and Hooi, Bryan},
  journal={arXiv preprint arXiv:2501.18492},
  year={2025}
}

@article{liu2025efficient,
  title={Efficient inference for large reasoning models: A survey},
  author={Liu, Yue and Wu, Jiaying and He, Yufei and Gong, Ruihan and Xia, Jun and Li, Liang and Gao, Hongcheng and Chen, Hongyu and Bi, Baolong and Zhang, Jiaheng and others},
  journal={arXiv preprint arXiv:2503.23077},
  year={2025}
}

@article{sui2025meta,
  title={Meta-reasoner: Dynamic guidance for optimized inference-time reasoning in large language models},
  author={Sui, Yuan and He, Yufei and Cao, Tri and Han, Simeng and Chen, Yulin and Hooi, Bryan},
  journal={arXiv preprint arXiv:2502.19918},
  year={2025}
}

@inproceedings{he2025enabling,
  title={Enabling Self-Improving Agents to Learn at Test Time With Human-In-The-Loop Guidance},
  author={He, Yufei and Li, Ruoyu and Chen, Alex and Liu, Yue and Chen, Yulin and Sui, Yuan and Chen, Cheng and Zhu, Yi and Luo, Luca and Yang, Frank and others},
  booktitle={Proceedings of the 2025 Conference on Empirical Methods in Natural Language Processing: Industry Track},
  pages={1625--1653},
  year={2025}
}

@article{he2025evotest,
  title={EvoTest: Evolutionary Test-Time Learning for Self-Improving Agentic Systems},
  author={He, Yufei and Liu, Juncheng and Liu, Yue and Li, Yibo and Cao, Tri and Hu, Zhiyuan and Xu, Xinxing and Hooi, Bryan},
  journal={arXiv preprint arXiv:2510.13220},
  year={2025}
}

@article{he2025evaluating,
  title={Evaluating the Paperclip Maximizer: Are RL-Based Language Models More Likely to Pursue Instrumental Goals?},
  author={He, Yufei and Li, Yuexin and Wu, Jiaying and Sui, Yuan and Chen, Yulin and Hooi, Bryan},
  journal={arXiv preprint arXiv:2502.12206},
  year={2025}
}

@article{chen2025can,
  title={Can Indirect Prompt Injection Attacks Be Detected and Removed?},
  author={Chen, Yulin and Li, Haoran and Sui, Yuan and He, Yufei and Liu, Yue and Song, Yangqiu and Hooi, Bryan},
  journal={arXiv preprint arXiv:2502.16580},
  year={2025}
}

@article{chen2025robustness,
  title={Robustness via Referencing: Defending against Prompt Injection Attacks by Referencing the Executed Instruction},
  author={Chen, Yulin and Li, Haoran and Sui, Yuan and Liu, Yue and He, Yufei and Song, Yangqiu and Hooi, Bryan},
  journal={arXiv preprint arXiv:2504.20472},
  year={2025}
}

@article{wang2025safety,
  title={Safety in large reasoning models: A survey},
  author={Wang, Cheng and Liu, Yue and Bi, Baolong and Zhang, Duzhen and Li, Zhong-Zhi and Ma, Yingwei and He, Yufei and Yu, Shengju and Li, Xinfeng and Fang, Junfeng and others},
  journal={arXiv preprint arXiv:2504.17704},
  year={2025}
}

@article{zhang2021scr,
  title={SCR: Training graph neural networks with consistency regularization},
  author={Zhang, Chenhui and He, Yufei and Cen, Yukuo and Hou, Zhenyu and Feng, Wenzheng and Dong, Yuxiao and Cheng, Xu and Cai, Hongyun and He, Feng and Tang, Jie},
  journal={arXiv preprint arXiv:2112.04319},
  year={2021}
}

@inproceedings{he2022sgkd,
  title={Sgkd: A scalable and effective knowledge distillation framework for graph representation learning},
  author={He, Yufei and Ma, Yao},
  booktitle={2022 IEEE International Conference on Data Mining Workshops (ICDMW)},
  pages={666--673},
  year={2022},
  organization={IEEE}
}

@article{cheng2025lps,
  title={LPS-GNN: Deploying Graph Neural Networks on Graphs with 100-Billion Edges},
  author={Cheng, Xu and Yao, Liang and He, Feng and Cen, Yukuo and He, Yufei and Zhang, Chenhui and Feng, Wenzheng and Cai, Hongyun and Tang, Jie},
  journal={arXiv preprint arXiv:2507.14570},
  year={2025}
}
